\documentclass[preprint,12pt,authoryear]{elsarticle}

\usepackage{amssymb}

\usepackage{graphicx}
\graphicspath{ {images/} }
\usepackage{caption}     
\usepackage{subcaption}

\usepackage{multirow}
\usepackage{array}
\newcolumntype{M}[1]{>{\centering\arraybackslash}m{#1}}

\usepackage{amsthm}
\usepackage{amsmath}
\usepackage{url}
\usepackage{algorithm}
\usepackage{algpseudocode}
\usepackage{algorithmicx}
\usepackage{lineno}
\usepackage{xcolor}

\algdef{SE}[DOWHILE]{Do}{doWhile}{\algorithmicdo}[1]{\algorithmicwhile\ #1}%
\algrenewcommand\algorithmicrequire{\textbf{Input:}}
\algrenewcommand\algorithmicensure{\textbf{Output:}}

\journal{arXiv}

\begin{document}

\begin{frontmatter}

\title{A 3D Multimodal Feature for Infrastructure Anomaly Detection}

\author[a]{Yixiong Jing}
\affiliation[a]{
    organization={
        Department of Engineering Science, 
        University of Oxford},
    addressline={Parks Road}, 
    city={Oxford},
    postcode={OX1 3PJ}, 
    state={Oxford},
    country={UK}}

\author[b]{Wei Lin}
\affiliation[b]{
    organization={
        Department of Geotechnical Engineering, 
        College of Civil Engineering, 
        Tongji University},
    addressline={1239 Siping Road}, 
    city={Shanghai},
    postcode={200092}, 
    state={Shanghai},
    country={China}}

\author[c]{Brian Sheil}
\affiliation[c]{
    organization={
        Construction Engineering, 
        University of Cambridge},
    addressline={Trumpington Street}, 
    city={Cambridge},
    postcode={CB2 1PZ}, 
    state={Cambridge},
    country={UK}}

\author[d]{Sinan Acikgoz}
\affiliation[d]{
    organization={
        Department of Engineering Science, 
        University of Oxford},
    addressline={Parks Road}, 
    city={Oxford},
    postcode={OX1 3PJ}, 
    state={Oxford},
    country={UK}}
    
\begin{abstract}
Ageing structures require periodic inspections to identify structural defects. Previous work has used geometric distortions to locate cracks in synthetic masonry bridge point clouds but has struggled to detect small cracks. To address this limitation, this study proposes a novel 3D multimodal feature, 3DMulti-FPFHI, that combines a customized Fast Point Feature Histogram (FPFH) with an intensity feature. This feature is integrated into the \textit{PatchCore} anomaly detection algorithm and evaluated through statistical and parametric analyses. The method is further evaluated using point clouds of a real masonry arch bridge and a full-scale experimental model of a concrete tunnel. Results show that the 3D intensity feature enhances inspection quality by improving crack detection; it also enables the identification of water ingress which introduces intensity anomalies. The 3DMulti-FPFHI outperforms FPFH and a state-of-the-art multimodal anomaly detection method. The potential of the method to address diverse infrastructure anomaly detection scenarios is highlighted by the minimal requirements for data compared to learning-based methods. The code and related point cloud dataset is available at \url{https://github.com/Jingyixiong/3D-Multi-FPFHI}.
\end{abstract}


\begin{keyword}
Anomaly detection \sep \textit{PatchCore} \sep point cloud \sep multimodal \sep crack detection \sep water patch detection \sep masonry bridge \sep tunnel
\end{keyword}

\end{frontmatter}

\section{Introduction}\label{sec1}
 
Critical civil infrastructure requires regular inspection to ensure safety and structural integrity. Over time, defects such as cracks and water patches may develop, which need to be detected and repaired actively to prevent further deterioration \citep{acikgoz2018sensing, 9678126}. Traditional inspection methods, which commonly rely on visual assessments, are labour-intensive, time-consuming, and often subjective, leading to variability in defect identification and classification \citep{BrackenburyDaniel2022AIIo}. With the increasing complexity and scale of modern infrastructure, there is a growing need for automated, data-driven inspection techniques that provide consistent and reliable assessments to support maintenance decisions.

Advances in deep learning (DL) provide opportunities to overcome these challenges with automated structural health monitoring of infrastructure. Recent work has facilitated automated classification, localization, and quantification of structural defects from image data \citep{cha2017deep, chen2017nb, cha2018autonomous, attard2019automatic, liu2019computer, li2020automatic, li2024comprehensive}. DL-based techniques have shown promise in detecting cracks in buildings \citep{perez2019deep, jiang2021building}, bridges \citep{dais2021automatic, hallee2021crack, loverdos2022automatic}, tunnels \citep{9678126, protopapadakis2019automatic}, and roads \citep{fan2020automatic}. For example, recent studies reported the successful detection of cracks with widths greater than $\leq$ 1 mm \citep{liao2022automatic, mohammadi2019non}. However, 2D image data acquisition is sensitive to illumination, which can adversely affect defect detection accuracy \citep{gupta2022image}. Furthermore, since 2D images lack geometric information, anomaly detection focuses only on surface anomalies. Given the strong correspondence between deformations and crack development, geometric information can help identify structurally significant cracks associated with geometric distortions. This approach may also help detect internal defects that, while not visible on the surface, cause noteworthy deformations \citep{acikgoz2017evaluation}.


Point clouds, acquired through laser scanning or photogrammetry, provide valuable data for detecting geometric distortions related to structural defects. Several studies have used point clouds to measure geometric deviations between structural components and predefined primitive geometries, such as planes or cylinders, to detect distortions \citep{pesci2011laser, liu2011lidar, ye2018mapping, dong2023pavement, del2022pavement, lin_novel_2023, lin_seg2tunnel_2024}. These distortions are typically identified by abrupt changes in elevation, normal vectors, and/or curvature. Alternatively, \cite{mohammadi2019non} detected damage on smooth concrete surfaces by identifying anomalies in geometric feature distributions. Other studies, such as \cite{lee2023new}, have extracted defect-induced geometric features from point cloud normal vectors, while \cite{stalowska2022crack} captured cracks on concrete surfaces by analyzing changes in laser beam intensity. Despite these advances, these methods rely on case-specific workflows, limiting their generalizability.

In recent work, \cite{jing4819836anomaly} adapted the \textit{PatchCore} \citep{roth2022towards} anomaly detection technique to detect cracks in masonry arch point clouds \citep{jing2023method}, achieving high accuracy and robustness. That method differentiates undeformed ('normal') point clouds from those with cracks ('anomalous') by extracting geometric features with Fast Point Feature Histograms (FPFH) \citep{rusu2009fast}. Cracks were detected by measuring discrepancies between paired points in 'normal' and 'anomalous' point clouds, effectively identifying both visible and non-visible cracks (e.g., intrados and extrados cracks in masonry arches). Unlike learning-based anomaly detection methods \citep{schlegl2019f, akcay2019ganomaly, beggel2020robust, nguyen2019anomaly, yi2020patch}, \textit{PatchCore} does not require extensive pre-training on normal datasets, making it practicable when data is scarce.

However, the anomaly detection method proposed by \cite{jing4819836anomaly} had three significant drawbacks:
\begin{enumerate}[(i)]
    \item It was observed that geometric features encoded in FPFH are not sufficiently sensitive to capture small cracks \citep{jing4819836anomaly}. Consequently, FPFH features were deemed limited to identifying defects with pronounced geometric distortions.
    \item Non-crack defects, such as water patches, are important for infrastructure defect detection. Such defects may not cause geometric distortions and are likely to be undetected in the original anomaly detection framework. 
    \item The method has only been validated on a synthetic arch dataset \citep{jing2023method}. While the synthetic dataset captured primary failure mechanisms, it does not explicitly simulate crack geometry. In other words, new surfaces created by the cracking were neglected.
\end{enumerate}

Drawbacks (i) and (ii) arise from using solely geometric features for anomaly detection. \cite{wang2023multimodal} proposed a multimodal Transformer-based model \citep{vaswani2017attention} to fuse two modalities (e.g., 3D geometry and 2D colour information) for improving anomaly detection accuracy. To enhance memory and computational efficiency, \cite{costanzino2024multimodal} used layer pruning to discard unnecessary layers from 2D and 3D feature extractors, while \cite{li2024collaborative} directly concatenated 2D and 3D feature maps for fusion. Despite achieving state-of-the-art (SOTA) performance on datasets such as MVTec 3D-AD \citep{bergmann2021mvtec}, current multimodal anomaly detection methods still rely heavily on Transformers for extracting the 3D feature and fusing different modalities, which are known to be memory intensive.

To avoid the heavy memory consumption of multimodal methods, \cite{cao2023complementary} proposed a complementary pseudo-multimodal feature (CPMF), which uses handcrafted descriptors that concatenate geometric features (extracted by FPFH) and 2D colour features (extracted by pre-trained 2D neural networks \citep{cao2022informative, wan2021industrial}). To obtain images for extracting 2D colour features, 3D point clouds are projected by a camera simulator with known intrinsic and extrinsic parameters in multiple views. The correspondence between pixels and points can therefore be determined which enables direct integration of 2D and 3D features. While CPMF demonstrates SOTA performance on MVTec 3D-AD, it may be constrained by the colour features extracted via pre-trained neural networks, which were not optimized for crack detection.

To overcome the aforementioned shortcomings, we introduce a multimodal feature, 3DMulti-FPFHI, which combines geometry and intensity features to provide robust anomaly detection. This approach leverages intensity information (obtained either from laser intensity or visible-light colour information), which can potentially enhance anomaly detection performance, especially regarding the detection of small structural cracks. Further, 3DMulti-FPFHI may enable the detection of non-geometric defects (such as water patches) that are characterized by intensity anomalies. To demonstrate the robustness of the proposed 3DMulti-FPFHI, CPMF is utilized as a baseline for comparisons in our synthetic data, supported by comprehensive statistical analysis. 

Several studies in the literature aimed to address drawback (iii) by improving synthetic simulators. A previous study by \cite{mohammadi2019non} introduced defects directly into the geometry, but neglected surface textures. \cite{narazaki2021synthetic} mapped concrete crack textures onto mesh surfaces, capturing visual but not geometric details. Recently, \cite{liu2024real3d} created the Real3D-AD benchmark, which manually adds anomalies to scanned high-resolution point clouds, though data collection and manipulation are non-trivial. By combining all the merits of previous studies, we propose a mechanically consistent approach for synthesizing the geometry of intrados cracks in the synthetic data in a fully automated manner.

Although previous studies \citep{jing2022segmentation, jing2024lightweight} have increased confidence that synthetic data can substitute real data for segmentation tasks, the effectiveness of such an approach for infrastructure anomaly detection remains an open question. To address this issue, the performance of anomaly detection techniques needs to be evaluated using real datasets. To this end, this paper explores the application of the framework to both a masonry arch bridge \citep{acikgoz2017evaluation} and a full-scale experimental model of a segmental concrete tunnel lining, where 'normal' and 'anomalous' data were clearly defined.

To summarize, the academic contributions of this work are as follows: (1) the introduction of 3DMulti-FPFHI as a novel multimodal feature for anomaly detection, benchmarked against a SOTA method CPMF; (2) the development of an enhanced synthetic dataset that bridges the gap between controlled experiments and real-world scenarios; and (3) the validation of the proposed framework on real-world datasets, demonstrating its generalizability for detecting defects across varied infrastructure types.

The paper is organized as follows. In Section 2, we detail the architecture of the anomaly detection framework, with a focus on the construction of the 3DMulti-FPFHI. Section 3 describes the synthetic and real-world datasets that we used to test the 3D Multi-FPFHI. Different from our previous work, \citep{jing4819836anomaly}, the new synthetic point clouds include intensity information and simulate new surfaces formed by cracks. Section 4 presents a comprehensive analysis of the proposed anomaly detection approach. The synthetic data are used to conduct feature analysis and sensitivity evaluations while the real-world datasets are used to assess the framework’s practical applicability in operational conditions. Finally, we discuss our findings and potential developments for future research in Section 5.

\begin{figure*}[t!]
    \centering
    \includegraphics[width=\linewidth]{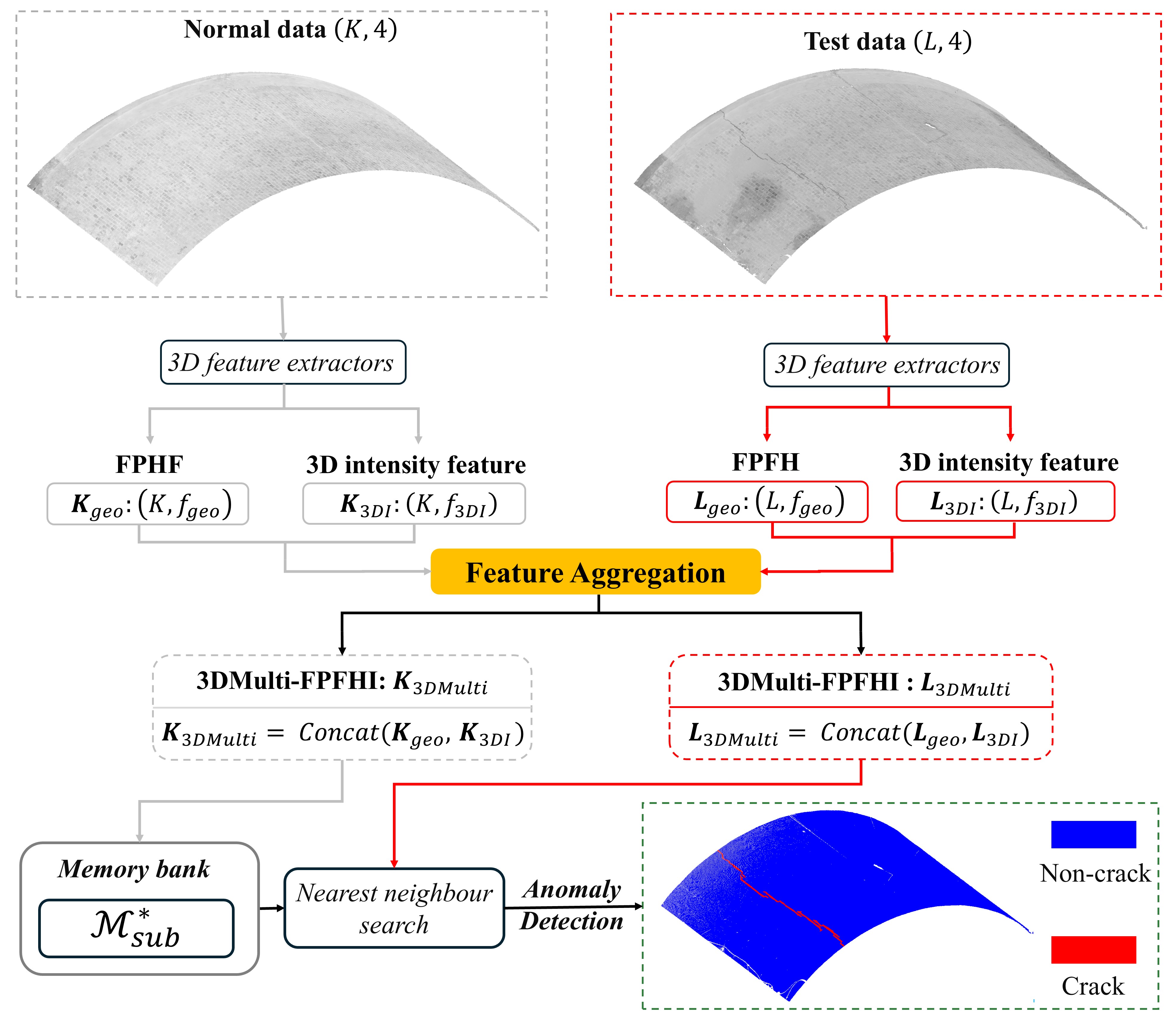}
    \caption{Overview of anomaly detection using \textit{PatchCore}. The input point cloud data consist of intensity values along with $x$, $y$, and $z$ coordinates. FPFH are extracted from both normal and test data (test data includes both normal and anomalous points) using point coordinates. The 3D intensity feature is derived from the intensity information of the point clouds. By concatenating FPFH and the 3D intensity feature, the proposed 3DMulti-FPFHI enhances anomaly detection accuracy. 3DMulti-FPFHI from the normal data are compressed into a memory bank and compared with test data features to detect anomalies, which are highlighted in red.}
    \label{fig: method_framework}
\end{figure*}

\section{Methodology}\label{sec2}

To explore the benefits of incorporating intensity information for anomaly detection, local intensity features are extracted using both 2D and 3D methods for comparison. The framework is shown in Figure \ref{fig: method_framework} and comprises three components: (1) geometric (e.g. FPFH) and 3D intensity feature extraction, and their aggregation in 3DMulti-FPFHI; (2) rendering of a 3D point cloud into images, followed by the extraction of 2D intensity features to generate CMPF; and (3) the utilization of \textit{PatchCore} for anomaly detection.

\subsection{FPFH and 3D intensity feature extraction and aggregation for 3DMulti-FPFHI}

FPFH \citep{rusu2009fast} is employed as a rotationally invariant geometric feature descriptor to capture local geometry. The computation of FPFH relies on the surface normal at each point in the point cloud. To obtain FPFH features, the ball query algorithm searches for neighbours around each point. For each neighbour-to-center point pair, a Darboux frame is defined, along with normal vectors to compute three intersection angles. These angles are embedded into three histograms, which are concatenated to describe normal, geodesic and torsional curvature. Geometric features from normal and test data are denoted as $\boldsymbol{K}_{geo}$ and $\boldsymbol{L}_{geo}$ respectively (see Figure \ref{fig: method_framework}). Implementation details can be found in \cite{jing4819836anomaly} and \cite{rusu2009fast}.

\begin{figure*}[t!]
    \centering
    \includegraphics[width=\linewidth]{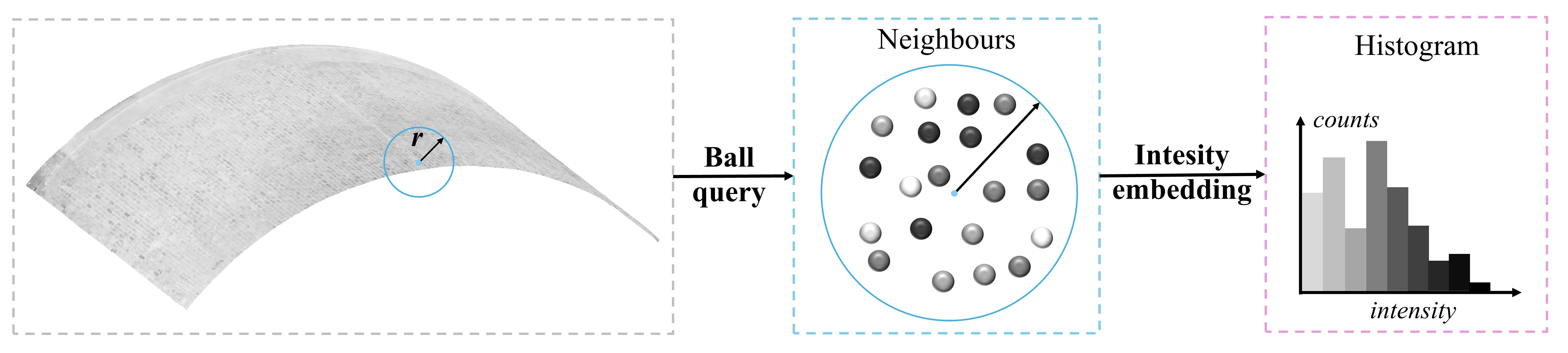}
    \caption{Computation of intensity histogram using FPFH. Intensity values of neighbours (defined by ball query with radius $r$) are incorporated into the histogram as intensity features.}
    \label{fig: intensity_fpfh}
\end{figure*}

For simplicity, it is assumed that intensity information for each point takes a scalar value (e.g., an infrared laser return amplitude or a visible light spectrum greyscale colour amplitude). To extract the 3D intensity feature, intensity values are encoded into a histogram, following a procedure analogous to FPFH. The method is illustrated in Figure \ref{fig: intensity_fpfh}: (1) Ball query is first adopted to obtain neighbours for each point within a radius $r$, and (2) local intensity features from these neighbours are then embedded into a histogram with a predefined bin size. The 3D intensity feature for normal and test data are represented as $\boldsymbol{K}_{3DI}$ and $\boldsymbol{L}_{3DI}$, respectively, as shown in Figure \ref{fig: method_framework}. Inspired by \cite{logoglu2016cospair}, this embedding computes the absolute relative values between the centre point and its neighbours,  reducing mismatches in the 3D intensity feature between normal and test data, which may occur due to changes in environmental conditions during data collection. This aspect is further discussed in Section 4.3. 

To enhance the representation of FPFH (e.g., $\boldsymbol{K}_{geo}$ and $\boldsymbol{L}_{geo}$), they are concatenated with 3D intensity feature $\boldsymbol{K}_{3DI}$ and $\boldsymbol{L}_{3DI}$ to create 3DMulti-FPFHI (e.g., $\boldsymbol{K}_{3DMulti}$ (normal data) and $\boldsymbol{L}_{3DMulti}$ (test data) as in Figure \ref{fig: method_framework}).

The original FPFH feature dimension is determined by the total number of bins across three histograms for three intersection angles, with each histogram containing 11 bins \citep{rusu2009fast}. However, experiments show that this bin count limits FPFH's ability to detect subtle geometric distortions due to its low resolution. To enhance anomaly detection accuracy, this study increases the bin count to 30 for both FPFH and the 3D intensity feature, providing finer resolution and better sensitivity.

\subsection{2D intensity feature extraction and aggregation for CMPF}

To leverage the strength of 2D deep learning methods, \cite{cao2023complementary} proposed extracting intensity features using 2D neural networks \citep{Lin_2017_ICCV}. These 2D features are concatenated with handcrafted 3D features (e.g., FPFH in their study) for the CPMF. CPMF was tested within the anomaly detection method \textit{PatchCore} \citep{roth2022towards}, which achieved SOTA performance on the MVTec 3D-AD dataset \citep{bergmann2021mvtec}. Thus, CPMF is used as a baseline to evaluate the performance of various 3DMulti-FPFHI features in this study.

\begin{figure*}[t!]
    \centering
    \includegraphics[width=\linewidth]{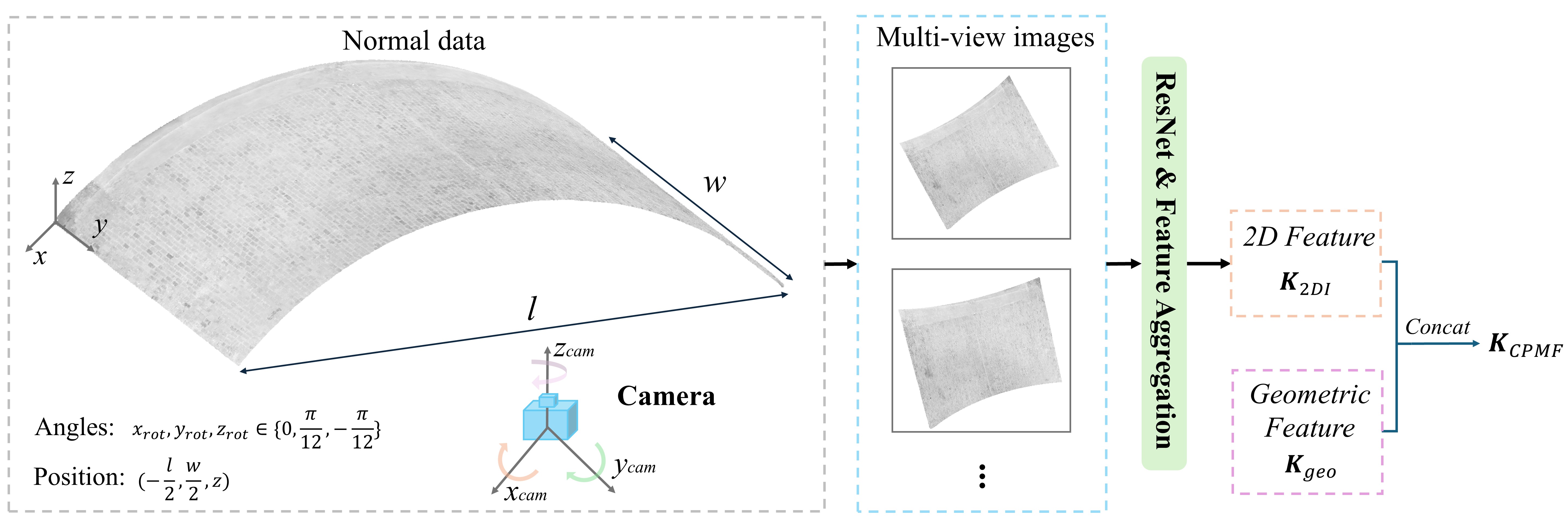}
    \caption{Computation of the pseudo-2D image feature on normal data. The point cloud intensity values are first projected to images using a synthetic camera (facing the positive $z$ direction in the initial position) with multiple viewing angles. Features are then extracted using ResNet50d \citep{he2016deep} and aggregated into the 2D feature $\boldsymbol{K}_{2DI}$.$\boldsymbol{K}_{2DI}$ is finally concatenated to $K_{geo}$ to form the CPMF.}
    \label{fig: pseudo 2d feature}
\end{figure*}

The computation of CPMF is illustrated in Figure \ref{fig: pseudo 2d feature}, where normal point cloud data are rendered into images using specified camera positions and rotation angles (e.g., $x_{rot}, y_{rot},$ and $z_{rot}$). These angles are sampled from a predefined set, resulting in 27 different orientations ($3\times3\times3$). The camera is centred in the arch, with its $z$ coordinate adjusted to cover the full structure. Initially, the camera faces the positive $z$ direction. With the specified rotation angles and position, a series of extrinsic matrices are determined for the camera. To capture the whole structure, the focal lengths of the camera for capturing synthetic and real arch point clouds are set to 700mm and 450mm respectively. The image size is set to $512\times512$ pixels, ensuring that the total pixel count approximates the number of points in the point cloud. The transformations between the point clouds and pixels of projected images can be straightforwardly obtained from the extrinsic and intrinsic matrices of the camera. 

This study uses the first three blocks from ResNet50d \citep{he2016deep}, with each feature map interpolated back to a $512\times512$ resolution and concatenated to form the 2D feature $\boldsymbol{K}_{2DI}\in \mathbb{R}^{K \times f_{2DI}}$, where $f_{2DI}$ equals 1792 (the sum of feature channels from the output of three blocks). $\boldsymbol{K}_{2DI}$ is normalized in feature dimension to ensure the same contribution as $\boldsymbol{K}_{geo}$ in the anomaly detection. CPMF for normal data, namely $\boldsymbol{K}_{CPMF}$ is obtained by concatenating $\boldsymbol{K}_{geo}$ with $\boldsymbol{K}_{2DI}$, as shown in Figure \ref{fig: intensity_fpfh}. The same procedure is applied to compute the test data (resulting in $\boldsymbol{L}_{CPMF}$), which is omitted for simplicity. Theoretical background and implementation details are provided in \cite{cao2023complementary}. 

\subsection{Anomaly detection method}

Anomaly detection is performed using the \textit{PatchCore} method \citep{roth2022towards}, which involves two key steps.  First, features extracted from normal data (e.g., $\boldsymbol{K}_{3DMulti}$ and $\boldsymbol{K}_{CPMF}$) are compressed using coreset subsampling \citep{sener2017active} to create a representative subset $\mathcal{M}_{sub}^*$ that approximates the original features. Coreset subsampling helps reduce redundancy in the memory bank, where normal features are stored (Figure \ref{fig: intensity_fpfh}). Secondly, test data features (e.g., $\boldsymbol{L}_{3DMulti}$ and $\boldsymbol{L}_{CPMF}$) are paired with features in $\mathcal{M}_{sub}^*$ using nearest neighbour search, where the Euclidean distances between feature pairs, $min\_dists$, are calculated. The individual values of $min\_dists$ are normalised with respect to the maximum value in the  $min\_dists$ set, and the values above the anomaly score $s$ are classified as anomalies. Further details are available in \cite{jing4819836anomaly} and the provided code.

\begin{figure*}[t]
    \centering
    \includegraphics[width=0.8\linewidth]{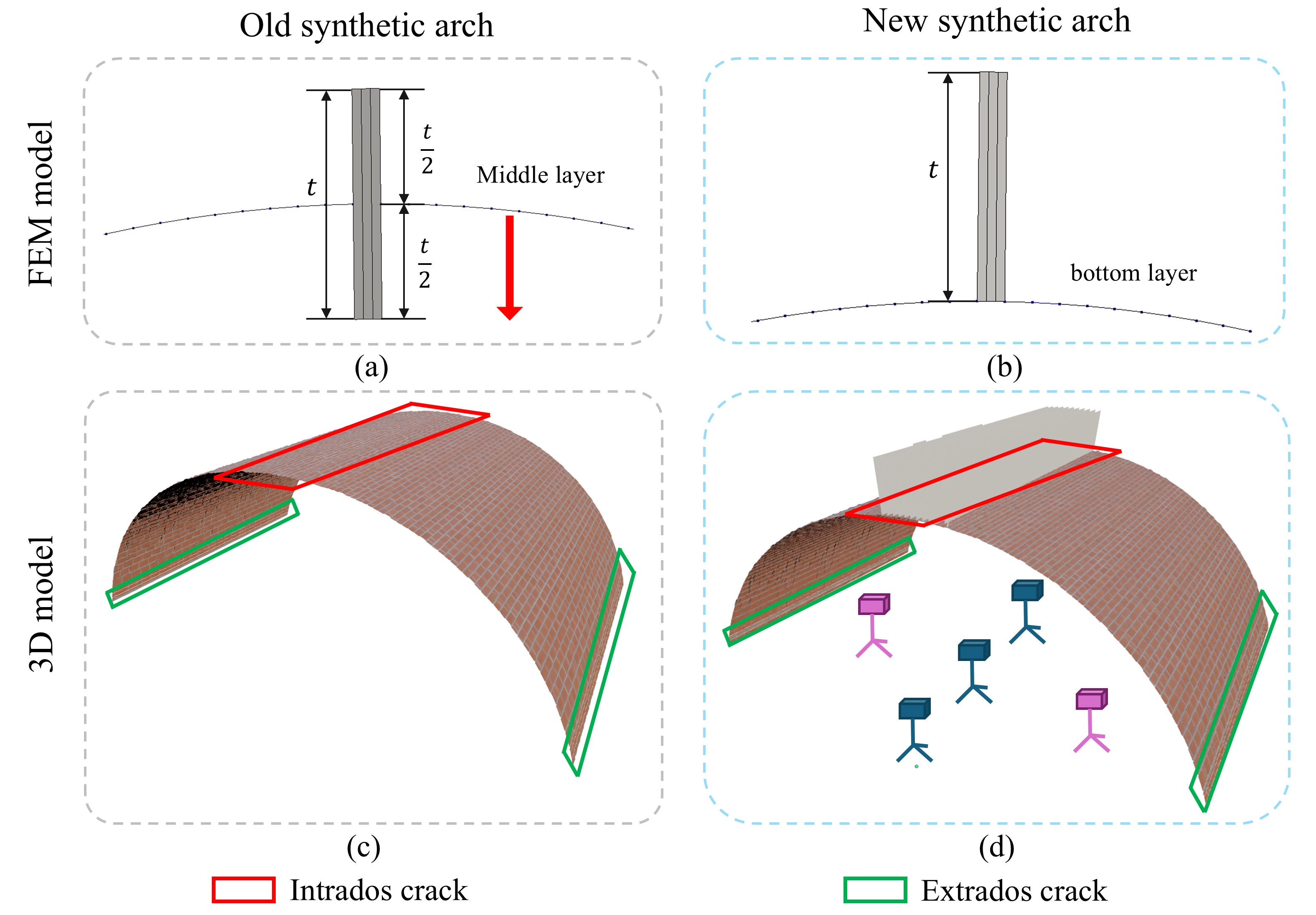}
    \caption{Modifications to the synthetic datasets, including updates to both the FEM model and its 3D representation in Blender. (a) and (b) show the layers where nodes are placed in the FEM model for the original and modified synthetic arches respectively. In the updated model, an eccentricity is applied by moving the node points to the bottom layer. (c) and (d) show the reconstructed 3D models in Blender for the original and modified arches, with intrados and extrados cracks highlighted in red and blue respectively. The new model now incorporates idealized intrados crack surfaces, points from which can be sampled by laser scanners.}
    \label{fig: syn_arch_data}
\end{figure*}

\begin{figure*}[t!]
    \centering
    \includegraphics[width=0.75\linewidth]{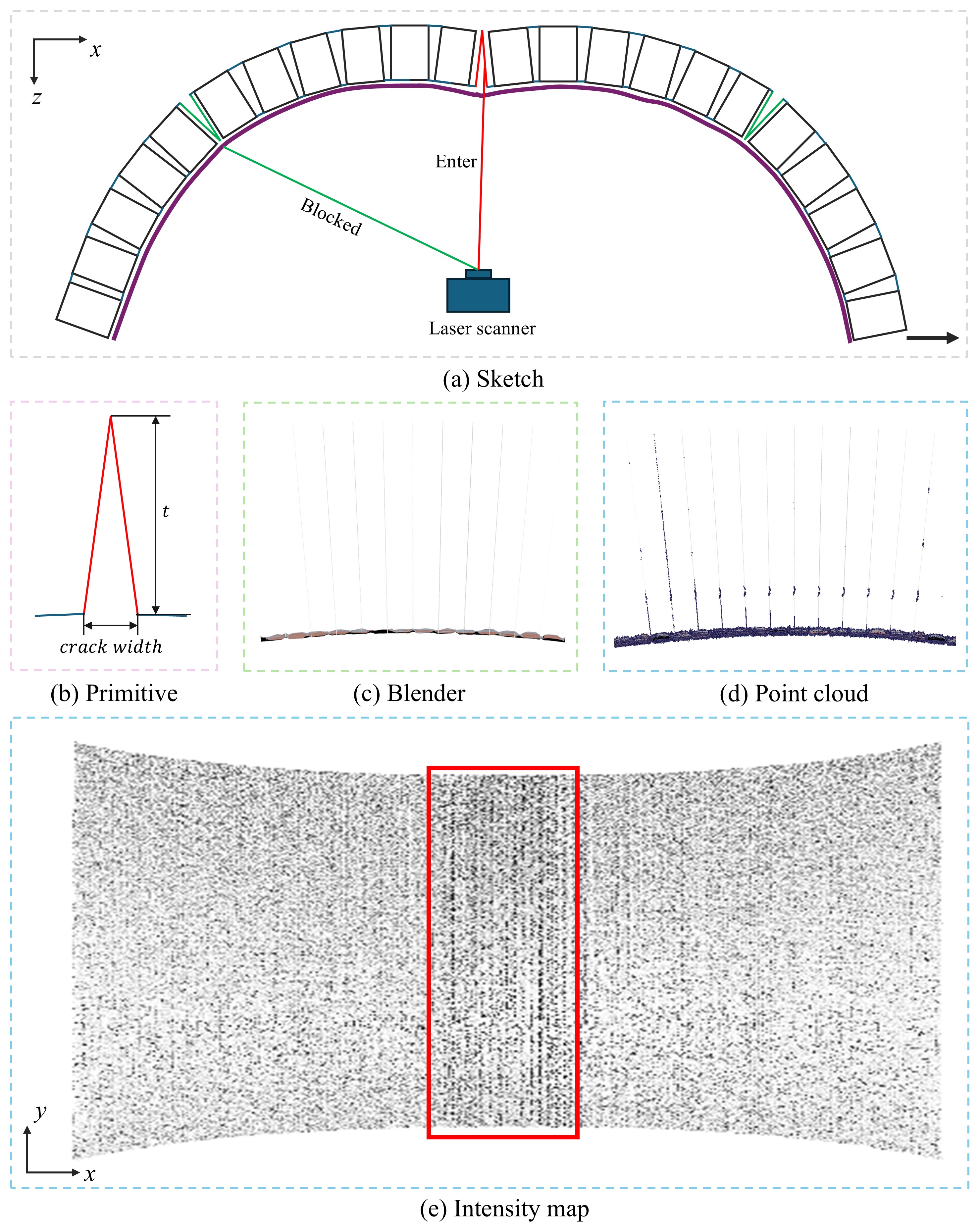}
    \caption{Integration of the intrados crack surfaces in the updated synthetic models. (a) shows a sketch comparing intrados (red colour) and extrados (green colour) cracks, highlighting how laser scanners may sample points from surfaces created by intrados cracks; (b) depicts the adopted idealized geometry of the intrados crack; (c) and (d) shows the reconstructed 3D model with intrados cracks in Blender and the synthetic point cloud, respectively. (e) represents the bottom view of the intensity map (greyscale) where the region of intrados cracks is shown in the red inset.}
    \label{fig: innerc_vis}
\end{figure*}

\section{Datasets}\label{sec3}

Three datasets are employed to evaluate and validate the proposed method: (1) a synthetic masonry arch dataset for statistical feature analysis and sensitivity evaluation; (2) a real masonry arch dataset; and (3) a full-scale experimental model of a concrete segmental tunnel.

\subsection{Synthetic masonry arch dataset}

The framework documented in \cite{jing4819836anomaly} is adopted to synthesize point clouds from finite element method (FEM) models of masonry arches. Details of the FEM modelling and point cloud generation process are provided in \cite{jing4819836anomaly} and are omitted here for brevity. However, modifications have been made to both the FEM model and the reconstructed 3D model in Blender for improved realism, as illustrated in Figure \ref{fig: syn_arch_data}a-d.

In the original synthetic arch FEM model (Figure \ref{fig: syn_arch_data}a), nodal displacements were extracted from integration points in the middle layer of the shell elements. However, these displacements are inconsistent with real-world scenarios since the laser scanner can only capture points on the intrados surface. To address this, the new FEM model prescribes an eccentricity of ${t}/{2}$, enabling the nodes to be located on the bottom layer of the shell, as illustrated in Figure \ref{fig: syn_arch_data}b, where ${t}$ is the thickness of the masonry units. The new model also distinguishes between intrados and extrados cracks, as shown in Figure \ref{fig: syn_arch_data}d. Extrados cracks can only be detected through geometric features since the surfaces they create are not visible on the intrados. In contrast, intrados cracks create surfaces which may be visible. 

To illustrate this aspect further, Figure \ref{fig: innerc_vis}a shows a deformed arch subjected to a support movement in the $x$ direction. The purple outline represents the deformed bottom arch surface. The laser scanner can sample points from the new surfaces created by intrados cracks (marked in red) but cannot sample points for extrados cracks (marked in green). In \citep{jing4819836anomaly}, the new surfaces created by intrados cracks were not considered (see Figure \ref{fig: syn_arch_data}c). However, in this study, a simple approach is developed to consider points from new surfaces created by intrados cracks. The idealized geometry of an intrados crack is shown in Figure \ref{fig: innerc_vis}b, where the front view represents an isosceles triangle of height $t$, and with a base width equal to the crack width on the bottom surface of the shell elements. Every time an intrados crack forms due to the separation of two shell elements, triangular crack surfaces are created. The reconstructed crack geometry and the corresponding point cloud for cracks in the crown area are shown in Figure \ref{fig: innerc_vis}c and d.

For simplicity, we assume uniform reflectivity across all structural surfaces (e.g., masonry, mortar, and inner crack surfaces). Consequently, variations in intensity between the inner crack surfaces and the rest regions are attributed to the geometry of the inner cracks. In the lidar simulator \citep{reitmann2021blainder}, synthesized laser beams undergo multiple reflections between the inner crack surfaces before returning to the sensor, resulting in lower intensity values. This phenomenon is visualized in the red inset of Figure \ref{fig: innerc_vis}e. The intensity map aligns with the experimental results of \citep{stalowska2022crack}, where crack regions exhibit lower intensity values compared to other surface regions.

\begin{figure*}[t!]
    \centering
    \includegraphics[width=\linewidth]{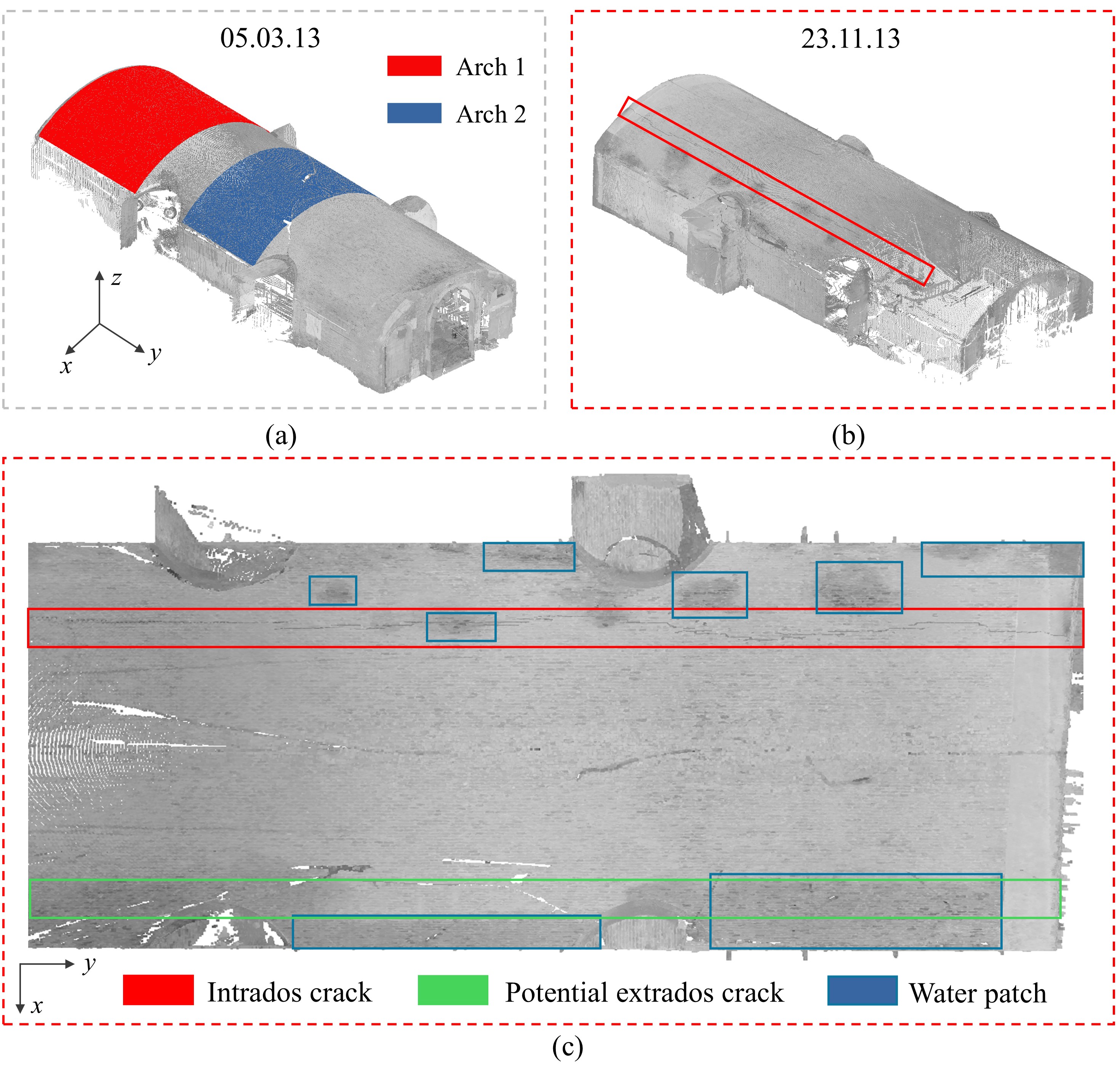}
    \caption{Visualization of the London Bridge Station point cloud for arch E57. (a) shows the point cloud collected on 05.03.13, highlighting segmented Arch 1 (blue) and Arch 2 (red) for anomaly detection. (b) indicates the point cloud collected on 23.11.13, showing a developed intrados crack in the red box. (c) represents the plan view of the point cloud on 23.11.13, with newly appeared water patches and a potential extrados crack highlighted by blue and green boxes, respectively.}
    \label{fig: real_arch_data}
\end{figure*}

The number of synthetic laser scanners has been increased from three to five in comparison to the previous setup in \cite{jing4819836anomaly}, (see Figure \ref{fig: syn_arch_data}d). In addition to the original three scanners (dark blue), two additional scanners (purple) are positioned on either side, spaced evenly along the direction orthogonal to the generatrix. This increased scanner count enhances the point density.

Surface roughness is defined by UV mapping brick and mortar depth profiles onto the masonry arch surfaces. Unlike the previous study \citep{jing4819836anomaly}, which only considered surface roughness for the $x$ support movement case, surface roughness is now included in all synthetic models to better reflect real masonry bridges. Surface roughness settings are consistent with those used in previous work \citep{jing4819836anomaly}. 

\subsection{Masonry arch dataset}

London Bridge Station, which was constructed during the 19th century, underwent a redevelopment focusing on the removal and then replacement of sections of its viaducts in 2013. This redevelopment included the construction of new piles and buttress walls supported by these piles \citep{acikgoz2017evaluation}. The most impactful phase, which introduced significant settlements, was the piling phase. During this phase, new piles were constructed under Arch E55. \cite{acikgoz2017evaluation} investigated the influence of these construction activities on the neighbouring Arch E57. Piling began on 31.01.13 and was completed by 16.08.13. The ensuing construction of buttress walls was completed by 21.11.13. 

\begin{figure*}[t]
    \centering
    \includegraphics[width=\linewidth]{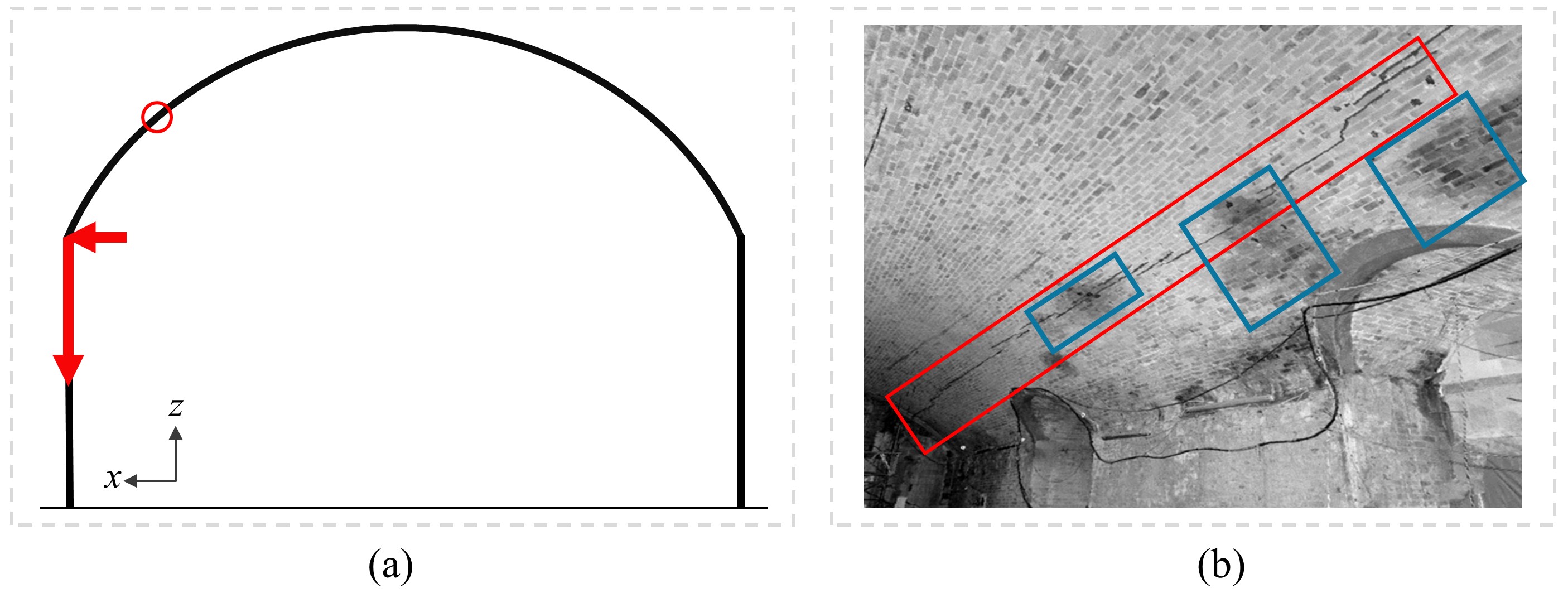}
    \caption{(a) Section view of Arch E57, highlighting settlement records indicated by red arrows. The intrados crack caused by the settlement is enclosed within a red circle. (b) Visualization of the intrados crack and water patches, marked with red and blue boxes, respectively.}
    \label{fig: real_arch_failure}
\end{figure*}

Point cloud data was collected using a Faro Focus X330 lidar scanner on three dates: 14.12.12, 05.03.13 and 23.11.13. These dates correspond to the periods when 0\%, 50\% and 100\% of the piles were constructed, respectively. However, the first scan on 14.12.12 was noisy and incomplete. Therefore, only the data collected on 05.03.13 and 23.11.13 are utilized in this study as shown in Figure \ref{fig: real_arch_data}a and b. Between these dates, an intrados crack (characterized by a span opening) of approximately 5mm was recorded and subsequently filled with mortar, as highlighted in the red box of Figure \ref{fig: real_arch_data}b. The intensity differences between the intrados crack (along the \textit{y}-axis) and the neighbouring non-crack regions are attributed to the large reflectivity differences between the filling mortar and the original materials. The intrados crack is evident in the plan view of the point cloud in Figure \ref{fig: real_arch_data}c, due to this intensity difference. Water patches were observed around the intrados crack, highlighted by blue boxes in Figure \ref{fig: real_arch_data}c.

In contrast, while no new crack formation is observed on the surface, a potential extrados crack (indicated indirectly by abundant nearby water patches) is highlighted in the green box in Figure \ref{fig: real_arch_data}c. However, it is not possible to definitively localize the extrados crack geometry, so a ground truth crack label was not attached to this location. 

\begin{figure*}[t]
    \centering
    \includegraphics[width=0.7\linewidth]{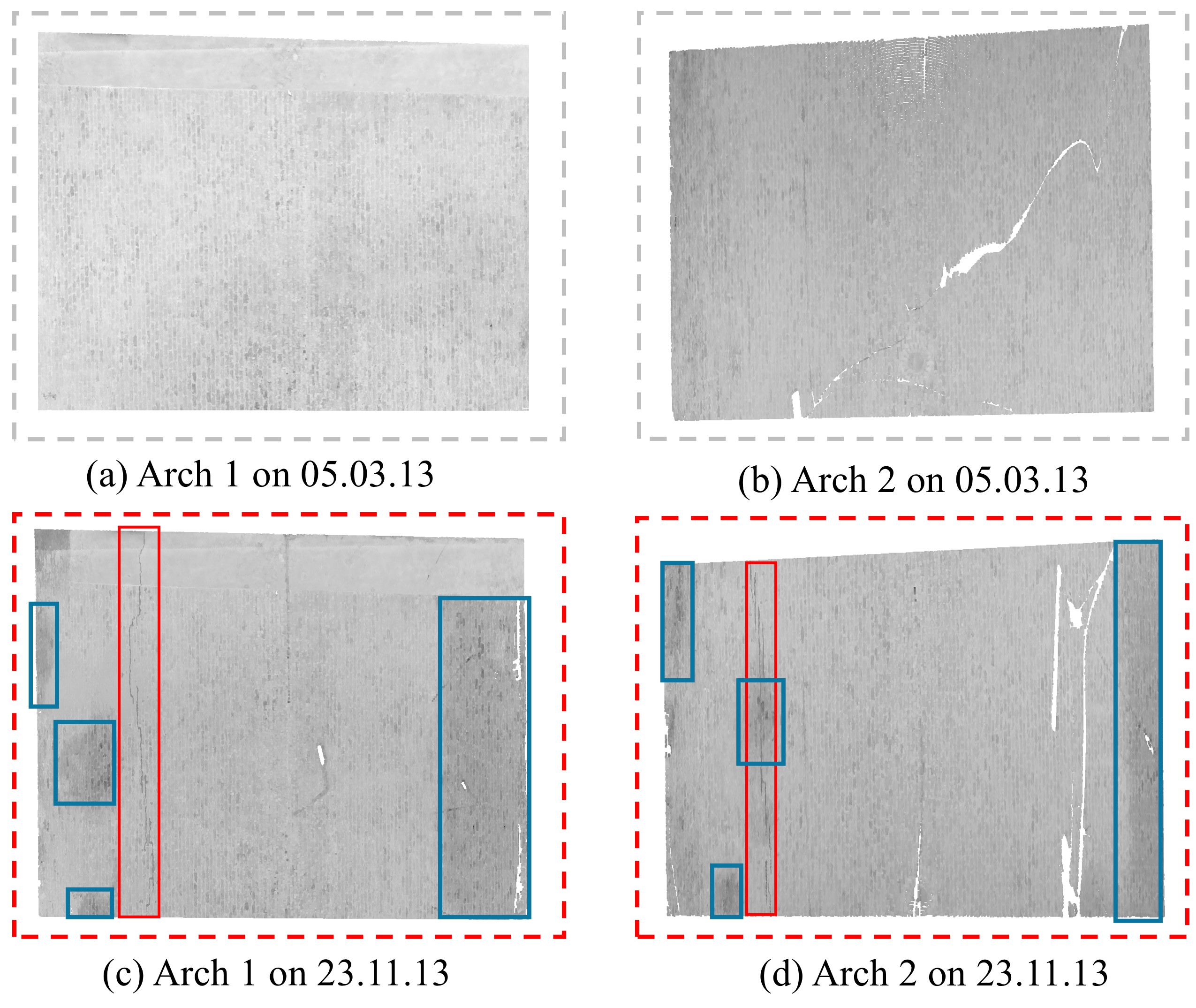}
    \caption{Top view of the segmented point clouds of arches 1 and 2, with intensity values mapped to greyscale. (a) and (b) represent the data collected on 05.03.13, showing no defects. (c) and (d) show the data from 23.11.13, highlighting an intrados crack (a red box) and water patches (blue boxes).}
    \label{fig: real_arch_separate}
\end{figure*}

Monitoring targets were installed to record support movements during the works. Significant vertical and lateral displacements were observed at the pier shared between Arches 55 and 57 on 23.11.13, as indicated by the red arrows in Figure \ref{fig: real_arch_failure}a. The intrados crack location is highlighted by a red circle. Figure \ref{fig: real_arch_failure}b visualizes the defects sustained by Arch E57 during the piling work, including the intrados crack and water patches.

\begin{table*}[t!]
    \centering
    \caption{Point numbers and the corresponding density measured on Arch 1 and Arch 2.}
    \renewcommand{\arraystretch}{1.2}
    \begin{tabular}{c|c|c|c|c}
         \hline
          {} & \multicolumn{2}{c|}{Arch 1} & \multicolumn{2}{c}{Arch 2} \\
         \cline{2-5}
         {} & 05.03.13 & 23.11.13 & 05.03.13 & 23.11.13 \\
         \hline
         Point number & {2,917,522} & {4,015,773} & {239,243} & {289,414} \\
         \hline
         Density(mm) & {3} & {3} & {13} & {11} \\
         \hline
    \end{tabular}
    \label{table: p_num_density}
\end{table*}

Anomaly detection was performed on the areas labelled Arch 1 and Arch 2 (highlighted in red and blue, respectively, in Figure \ref{fig: real_arch_data}a). These two arches were segmented from the full point cloud and analyzed separately, as shown in Figure \ref{fig: real_arch_separate}a-d. The point numbers and corresponding densities are shown in Table \ref{table: p_num_density}, where density is measured by averaging the nearest neighbour distances in the point clouds. Although this study focuses primarily on crack detection, water patches, were also defined as anomalies to demonstrate the potential benefits of integrating the intensity feature. Further details on the case study of Arch E57 can be found in \cite{acikgoz2017evaluation}.

\begin{figure*}[t!]
    \centering
    \includegraphics[width=0.75\linewidth]{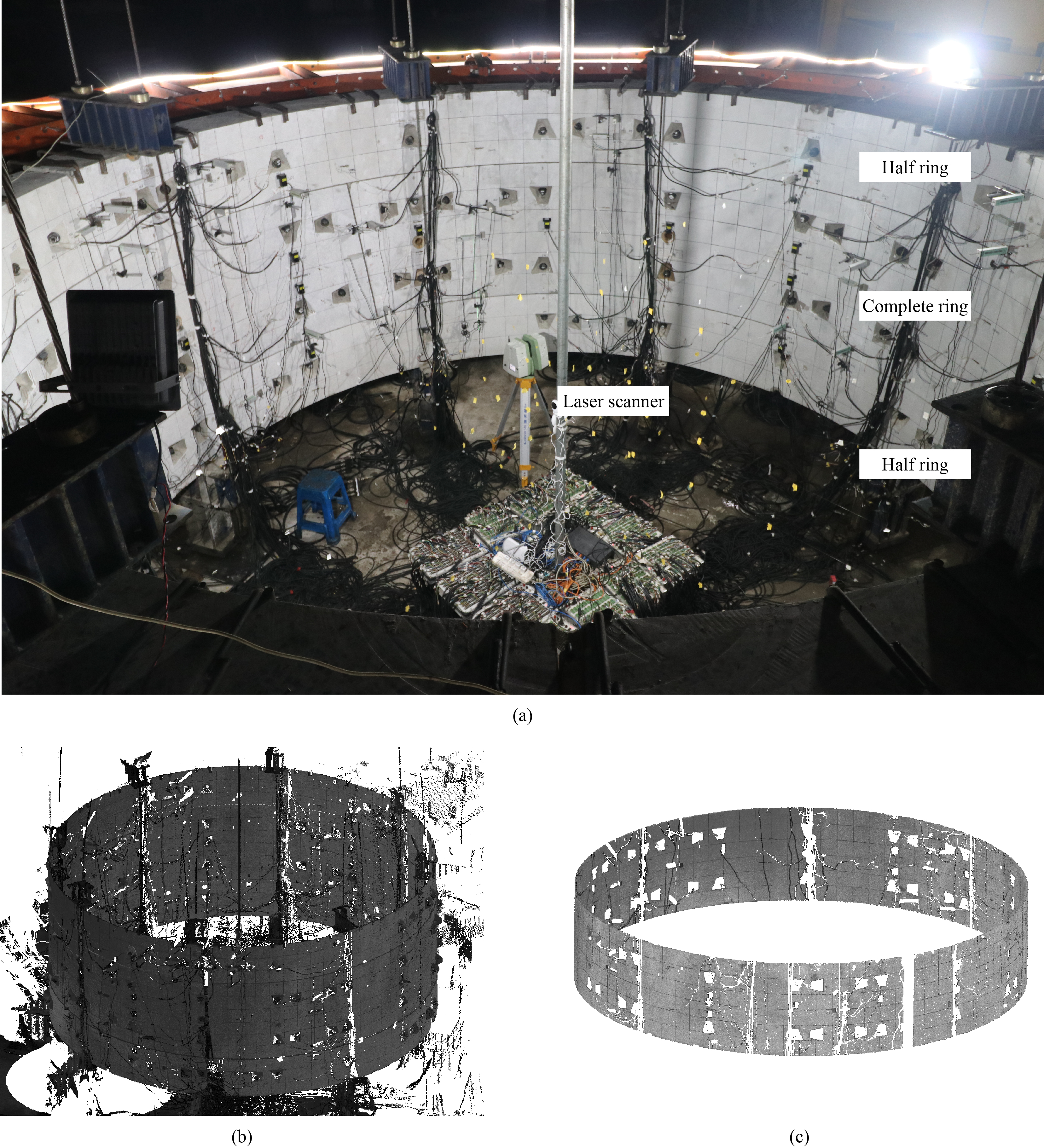}
    \caption{Visualization of the full-scale physical experiment and the resulting point clouds. (a) shows the experimental setup, including the laser scanner and tunnel lining rings. (b) shows the raw registered point cloud while (c) shows the processed (segmented and cleaned) point cloud of the middle complete ring.}
    \label{fig: experiment_layout}
\end{figure*}

\begin{figure*}[t]
    \centering
    \includegraphics[width=\linewidth]{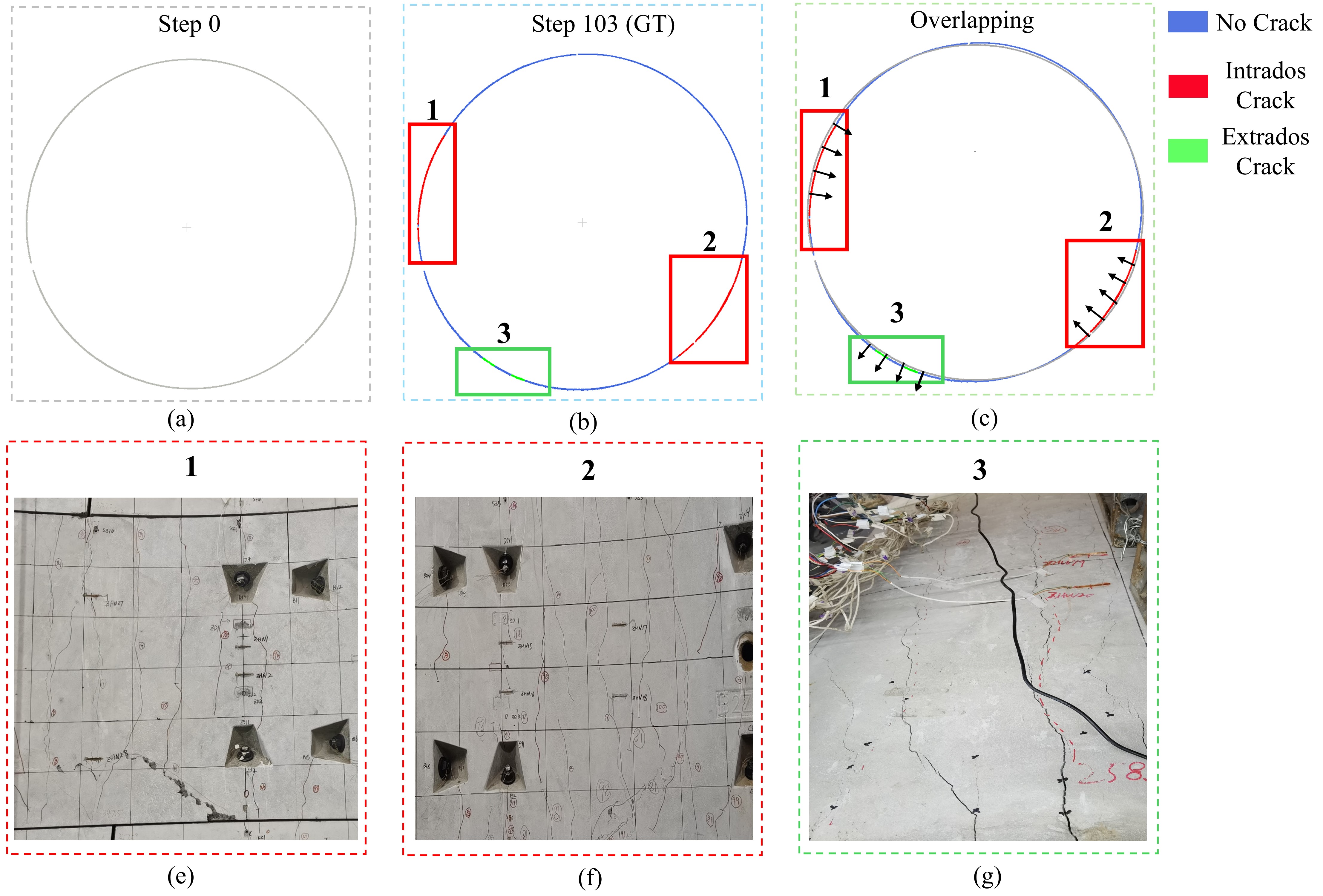}
    \caption{Visualization of the geometric distortion of the real tunnel in bird view and corresponding crack photos. (a) visualizes the unloaded tunnel; (b) ground truth on loading step 103 (blue, red, and green indicate the no crack, intrados crack, and extrados crack labels); (c) is retrieved by overlapping (b) and (c) where deformations are indicated by black arrows; (e), (f), and (g) show photos of cracks indexed as \textbf{1}, \textbf{2}, and \textbf{3} respectively.}
    \label{fig: bird_tunnel}
\end{figure*}

\subsection{Concrete segmental tunnel dataset}

Point clouds were acquired during a full-scale experiment on a segmental concrete tunnel lining. The experimental structure consists of one complete (e.g. full-length in the tunnel longitudinal direction) lining ring and two half-rings (e.g. half-length in the tunnel longitudinal direction), as shown in Figure \ref{fig: experiment_layout}a and b. The complete ring is stagger-jointed and the half-ring on either side replicates the in-situ boundary conditions. Only point clouds from the complete ring (Figure \ref{fig: experiment_layout}c) are used for anomaly detection. The lining consists of six concrete segments per ring, with an inner diameter of 2750 mm, a thickness (radial direction) of 350 mm, and a length (longitudinal direction) of 1200 mm. 

In the experiment, external loads were applied to the lining by a set of servo-hydraulic jacks positioned concentrically around the rings to simulate earth pressure. Loading continued until structural failure, meaning the dataset captures a wide range of conditions — from intact to severely damaged. There were a total of 103 loading steps.

The point clouds were scanned using a terrestrial laser scanner (Leica ScanStation C10). Given the lack of occlusions, a single scan was used to capture the whole structure in each loading step. In addition to the geometry information, laser intensity information was also recorded. Point clouds were acquired at loading steps 0, 20, 25, 76, 89, 96 and 103. 

The point clouds of the complete ring were manually segmented and non-lining points were removed. The resulting clean point clouds of the complete ring have \textasciitilde 3,000,000 points each and the average distance between neighbouring points is 2 mm. Cracks were recorded after each pre-defined loading step, allowing anomaly detection results to be validated against visual observations. 

We visualize the point cloud corresponding to step 0 (e.g., the reference geometry) and step 103 with cross-sections in Figure \ref{fig: bird_tunnel}a and b. Regions containing intrados cracks (indices 1 and 2) are highlighted in red boxes in Figure \ref{fig: bird_tunnel}b, and were directly captured by the laser scanner. In contrast, extrados cracks were not directly captured by the laser scanner. An example extrados crack (shown with index 3) is highlighted with a green box in Figure \ref{fig: bird_tunnel}. By overlaying Figure \ref{fig: bird_tunnel}a and b, the resultant Figure \ref{fig: bird_tunnel}c illustrates the deformation of tunnel rings in the crack regions under the applied force. Intrados cracks are characterised by inward bulging, while extrados cracks exhibit outward bulging, indicated by black arrows. Figure \ref{fig: bird_tunnel}e-g provide photographs of the cracks.

Crack labels were needed to enable an evaluation of the performance of the algorithm. The surface of the concrete was marked with a 20cm square grid as shown on the concrete surface of Figure \ref{fig: experiment_layout}a. If a crack is observed within this grid,  all points in the point cloud in this region are associated with a crack label. Then due to the 2D nature of the experimental setup, it was assumed that the cracks extended uniformly along the length of the ring. The labelling was performed for each acquired point cloud, by identifying the newly added cracks (compared to the last scanned loading step) as shown with black mark lines in Figure \ref{fig: bird_tunnel}e-g. 

\section{Experiments and results}\label{sec4}

\subsection{Implementation details and metrics}

To improve computational efficiency and achieve uniform density, point clouds from different datasets are subsampled via voxel downsampling. The synthetic and real arch point clouds are downsampled to a voxel size of 0.02m, while the tunnel point clouds are downsampled to 0.01m to retain surface detail. Ball query with a radius of 0.12 m is used to identify neighbouring points to accurately and robustly estimate the surface normals. A 1m radius is used for FPFH feature computation to capture the local geometry of synthetic and real masonry arch, whilst the radius is adjusted to 0.6m for the tunnel dataset. The size of $\mathcal{M}_{sub}^*$ (used for storing normal features) is set to 4000.

The intrinsic and extrinsic matrix settings for the 2D-to-3D projection are included in the provided code. CPMF is computed and compared across different feature types using synthetic data only, as our results and statistical analysis clearly highlight the limitations of CPMF when directly applied to infrastructure anomaly detection, which will be elaborated on in Section 4.2. 

Experiments are conducted on an AMD Ryzen 9 5900 CPU with 128GB memory (for FPFH computation) and RTX 3080 with 10GB memory (for feature map computation in images).  Since the relationship between geometric anomalies and crack width is non-linear, a single optimal anomaly score $s$ cannot be defined for binary classification across all datasets. Therefore it is difficult to use traditional performance metrics (e.g., $F1\textit{-}score$) to quantify the performance of the features for anomaly detection. To investigate this aspect, Section 4.2.2 introduces statistical analysis approaches for a more comprehensive evaluation of feature types and their influence on anomaly detection results. The conclusions drawn from this analysis can be generalized to other tests. However, $F1\textit{-}score$ with varied thresholds (e.g., $0.3*s$ and $0.5*s$) are still provided in Sections 4.2 to 4.4 for comparing the performance of different features.

\begin{figure*}[t!]
    \centering
    \includegraphics[width=\linewidth]{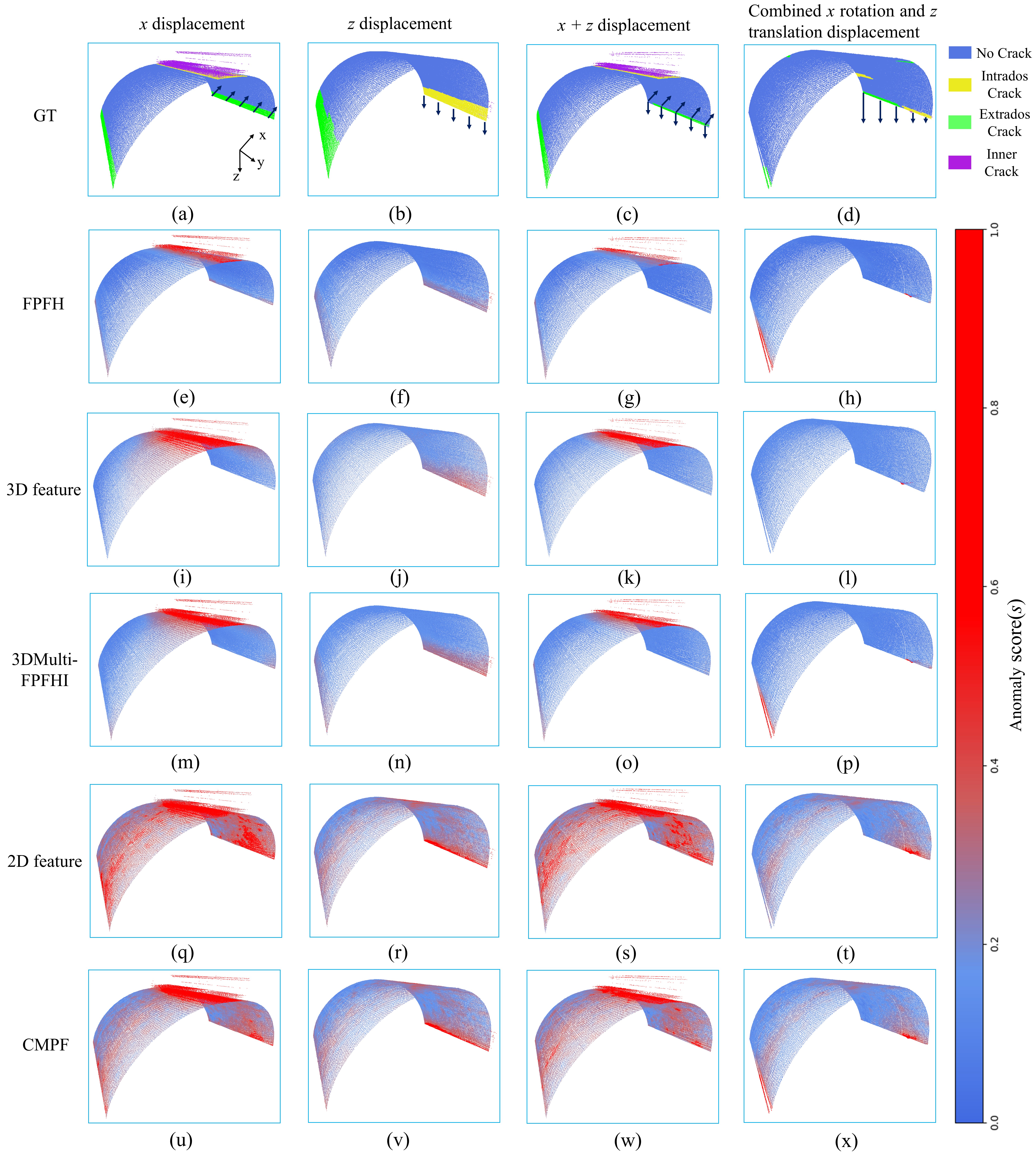}
    \caption{Anomaly detection results on the masonry arch for four different support movement cases. (a)-(d) show the ground truth where blue, yellow, green, and purple colours represent non defect, intrados crack, extrados crack and inner crack correspondingly; (e)-(h), (i)-(l), (m)-(p), (q)-(t), and (u)-(x) display heatmaps representing the anomaly score for each point, using different feature types.}
    \label{fig: syn_arch_result_0}
\end{figure*}

\subsection{Results for the synthetic arch}

\subsubsection{Anomaly detection under different support movement cases}

The masonry arch FEM model is subjected to four distinct support movements: $x$, $z$, $x+z$, and a combination of $x$ rotation and $z$ translation, as shown in Figure \ref{fig: syn_arch_result_0}a-d respectively. Although \textit{PatchCore} does not differentiate crack types, points are categorised as 'no crack', 'intrados crack', 'extrados crack', and 'inner crack' (represented as blue, yellow, green, and purple respectively in Figure \ref{fig: syn_arch_result_0}a-d). Specifically, 'inner crack' refers to points on the intrados cracks where points were sampled from the new surfaces created by cracks. The sub-classification of crack types will later assist in analyzing how different crack geometries influence anomaly detection, allowing for more in-depth statistical analysis.  

Results are visualized using heatmaps in Figure \ref{fig: syn_arch_result_0}e-x, where the anomaly scale is in the range [0, 1]. Points with an anomaly score above 0.5 are shown in red, whilst the remaining points are illustrated on a blue-to-red scale, representing increasing geometric anomalies. The threshold of 0.5 was chosen as it highlights extrados cracks more clearly; these cracks are often associated with low anomaly scores as their geometric distortions are less pronounced than other cracks. 

The performance of the anomaly detection method is compared across five different feature types in the synthetic dataset. Figure \ref{fig: syn_arch_result_0}e-h illustrate the detection performance using FPFH, where both intrados and extrados cracks are identified using geometric information. Then, Figure \ref{fig: syn_arch_result_0}i-l provide the anomaly detection results using only the 3D intensity feature. Compared to FPFH, the anomaly scores associated with the 'intrados crack' showed greater differentiation from normal data for the 3D intensity feature. The reason for this relates to the intensity anomalies introduced by new crack surfaces around the intrados cracks. However, the 3D intensity feature failed to detect extrados cracks, as intensity values remain invariant in regions where no new crack surfaces are created.

Additionally, the 3D intensity feature introduces diffusion of anomalies for intrados cracks, where anomalies spread from the cracks to adjacent normal regions with gradually decreasing anomaly scores, as evident in Figure \ref{fig: syn_arch_result_0}i and k. The diffusion is caused by the large size of the selected ball query radius (e.g. 1m in the synthetic arch dataset). The diffusion occurs for both intrados and extrados cracks, potentially affecting the precision of localizing anomalies.

\begin{table*}[t!]
\centering
\caption{$F1\textit{-}score$ for the four different support movement cases of the masonry arch with thresholds of $0.3*s$ and $0.5*s$. Red text indicates the best metrics whereas blue text denotes the second-best metrics.}
\label{table: f1 4 moves}
\renewcommand{\arraystretch}{1.2}
\begin{tabular} 
 {M{4cm}|M{1.2cm}|M{1.5cm}|M{1.5cm}|M{1.3cm}|M{1.3cm}}
 \hline
 {} & FPFH & 3D feature & 3DMulti-FPFHI & 2D feature & CMPF\\
 \hline
 $x(0.3*s)$ & \textcolor{red}{0.730} & 0.596 & \textcolor{blue}{0.676} & 0.430 & 0.430 \\
 $x(0.5*s)$ & 0.240 & 0.598 & \textcolor{red}{0.673} & 0.444 & \textcolor{blue}{0.669} \\
 \hline
 $z(0.3*s)$ & 0.120 & \textcolor{blue}{0.453} & \textcolor{red}{0.505} & 0.298 & 0.299 \\
 $z(0.5*s)$ & 0.090 & 0.097 & 0.102 & \textcolor{blue}{0.486} & \textcolor{red}{0.531} \\
 \hline 
 $x\,+\,z(0.3*s)$  & 0.313 & \textcolor{blue}{0.639} & \textcolor{red}{0.678} & 0.316 & 0.316 \\
 $x\,+\,z(0.5*s)$  & 0.184 & \textcolor{red}{0.680} & \textcolor{blue}{0.639} & 0.330 & 0.441 \\
 \hline
 $x\, rot\,+\,z\, trans(0.3*s)$  & \textcolor{blue}{0.310} & 0.213 & \textcolor{red}{0.330} & 0.060 & 0.072\\
 $x\, rot\,+\,z\, trans(0.5*s)$  & 0.302 & 0.182 & 0.318 & \textcolor{blue}{0.368} & \textcolor{red}{0.427} \\
 \hline
\end{tabular}
\end{table*}

Detection results using the 3DMulti-FPFHI are presented in Figure \ref{fig: syn_arch_result_0}m-p. 3DMulti-FPFHI preserves the advantages of the 3D intensity feature in detecting intrados cracks while effectively capturing extrados cracks by integrating FPFH. As shown in Figure \ref{fig: syn_arch_result_0}m and o, the diffusion effect observed with the 3D intensity feature is mitigated in 3DMulti-FPFHI. This improvement is achieved by FPFH balancing the contribution of the 3D intensity feature, suppressing the anomaly scores in diffused regions.

Figure \ref{fig: syn_arch_result_0}q-t illustrate the performance when using the 2D intensity feature. While the 2D intensity feature can roughly identify the locations of intrados cracks by detecting intensity anomalies in 2D pixels, anomaly detection remains imprecise, even when combined with FPFH to form CMPF, as shown in Figure \ref{fig: syn_arch_result_0}u-x. The 2D intensity feature struggles to capture the fine details of cracks. One potential reason for this relates to the ImageNet training data \citep{5206848}, which lacks images representing infrastructure anomalies (e.g., cracks). 

Anomaly detection results across all five features are quantified in Table \ref{table: f1 4 moves} using the $F1\textit{-}score$ metric. 3DMulti-FPFHI demonstrates superior performance compared to other feature types in most cases. Notably, the 2D intensity feature and CMPF achieve higher performance for the $x\, rot\,+\,z\, trans$ case when the threshold is 0.5$s$. However, results shown in Figure \ref{fig: syn_arch_result_0}q-t and u-x clearly indicate CMPF's inability to localize cracks; $F1\textit{-}score$ metric does not capture this deficiency.

\begin{figure*}[t!]
    \centering
    \includegraphics[width=\linewidth]{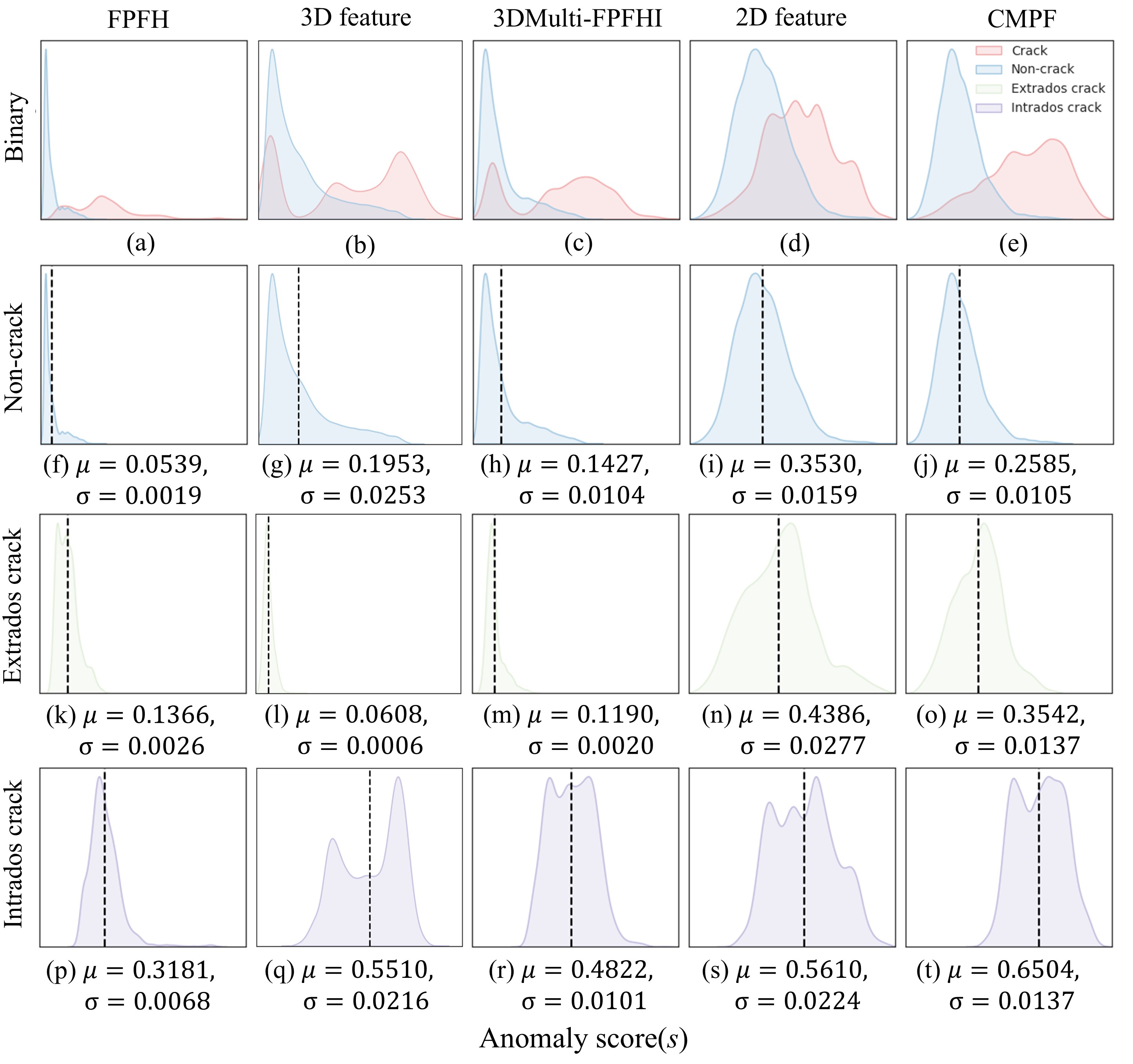}
    \caption{Distributions of $min\_dists$ for the $x$ displacement case. (a)-(e) represents distributions of 'non-crack' (blue) and 'crack' (red) points for five different feature types. (f)-(j) show distributions of 'non-crack' points together with their corresponding mean and variance. The 'crack' label is further subdivided into 'extrados crack' (green) and 'intrados crack' (purple) shown in (k)-(o) and (p)-(t) respectively to discriminate the influence of feature type on statistical metrics of $min\_dists$.}
    \label{fig: feature_hist}
\end{figure*}

\subsubsection{Statistical analysis for the $x$-direction support movement case}

To investigate further, the values of $min\_dists$ for the five different feature types are approximated using Kernel Density Estimation (KDE) \citep{wkeglarczyk2018kernel} and represented with continuous probability density functions. In these distributions, the horizontal axis denotes the $min\_dists$ values, while the vertical axis represents the estimated probability density. The KDE distributions for the '$x$-displacement' case are illustrated in Figure \ref{fig: feature_hist}a-t using FPFH, 3D intensity feature, 3DMulti-FPFHI, 2D intensity feature, and CMPF respectively. For an initial high-level analysis, 'intrados crack', 'extrados crack', and 'inner crack' are grouped under the 'crack' label. Further crack analyses fuse 'intrados crack' and 'inner crack' labels under 'intrados crack' (Figure \ref{fig: innerc_vis}d) for simplicity. 

The distribution of 'non-crack' and 'crack' labels are shown in Figure \ref{fig: feature_hist}a-e, where overlapping regions indicate the regions where the feature is unable to differentiate between the normal and anomalous data. When a threshold is applied, 'non-crack' points on the right side of the threshold are misclassified as 'crack', whilst 'crack' points on the left side of the threshold are erroneously identified as 'non-crack'. These results demonstrate that the handcrafted features (e.g., FPFH, 3D intensity feature, and 3DMulti-FPFHI) outperform the learning-based 2D intensity feature and CMPF by providing a clearer boundary between 'non-crack' and 'crack' distributions.
 
To quantitatively assess the influence of feature type on model performance, the 'non-crack' distributions are examined, where the mean value $\mu$ is indicated by a black vertical dotted line in Figure \ref{fig: feature_hist}f-j. Ideally, $\mu$ would approach 0 with a small standard deviation $\sigma$, indicating precise matching of 'non-crack' points before and after support movements. However, due to non-rigid displacements and measurement errors, the anomaly scores of normal points assume non-zero values. This is best illustrated using data from regions near cracks which also experience geometric distortions; points from these regions likely end up on the right tail of the distribution in Figure \ref{fig: feature_hist}f. Points in this region deviate from a normal distribution and increase the standard deviation $\sigma$. Compared to FPFH (Figure \ref{fig: feature_hist}f), the 3D intensity feature and 3DMulti-FPFHI are associated with a larger value of $\mu$; this is due to the introduction of intensity features. However, their average values are significantly less than the ones for 2D intensity feature and CMPF, explaining why the latter features misidentified areas in Figure \ref{fig: syn_arch_result_0}q and u. Notably, combining intensity features with FPFH (3DMulti-FPFHI and CMPF) reduces $\mu$ and $\sigma$ for 'non-crack' distributions, which indicates an important benefit of aggregating two modalities.

The 'extrados crack' and 'intrados crack' distributions are shown in Figure \ref{fig: feature_hist}k-o and p-t, respectively. A comparison of the $\mu$ values for FPFH (Figure \ref{fig: feature_hist}k and p) reveals that 'extrados cracks' in the examined scenario ($\mu = 0.1366$) are associated with smaller geometric distortions than 'intrados cracks' ($\mu = 0.3181$). Therefore, 'extrados cracks' are more difficult to detect with FPFH, as their $\mu$ is closer to the 'non-crack' distribution ($\mu = 0.0539$, Figure \ref{fig: feature_hist}f).

For the 3D intensity feature, $\mu$ and $\sigma$ are higher for the 'non-crack' distribution (Figure \ref{fig: feature_hist}g) compared to FPFH, leading to increased overlap between 'crack' and 'non-crack' regions and reduced localization precision. Furthermore, the 'extrados crack' distribution shows smaller $\mu$ and $\sigma$ compared to the 'non-crack' distribution (Figure \ref{fig: feature_hist}g and l), making it infeasible to differentiate between them, which corresponds to the results shown in Figure \ref{fig: syn_arch_result_0}i-j. In contrast, the 'intrados crack' distribution demonstrates a significant difference in $\mu$ from the 'non-crack' distribution, enabling intrados crack detection (Figure \ref{fig: feature_hist}q). However, the increased $\sigma$ for 'intrados crack' distributions (e.g., $\sigma = 0.068$ for FPFH and $\sigma = 0.0216$ for the 3D intensity feature corresponding to Figure \ref{fig: feature_hist}p and q) highlights the diffusion effect, where anomaly scores spread from cracks to adjacent points with lower values.

The 3DMulti-FPFHI combines the strengths of FPFH and the 3D intensity feature. While $\mu$ and $\sigma$ remain higher than FPFH for the 'non-crack' distribution, the differences are minor due to FPFH mitigating the diffusion effect of the 3D intensity feature (Figure \ref{fig: feature_hist}h and m). For 'intrados cracks,' $\mu$ significantly increases with 3DMulti-FPFHI ($\mu = 0.4822$, Figure \ref{fig: syn_arch_result_0}r) compared to FPFH alone, indicating advancements of considering the additional intensity modality. 

Figure \ref{fig: feature_hist}n, o, s, and t show that the addition of 2D intensity feature and CMPF lead to large $\sigma$ for both 'extrados crack' ($\sigma=0.0277$ and  $\sigma=0.0137$ respectively) and 'intrados crack' distributions ($\sigma=0.0224$ and $\sigma=0.0137$ respectively). The distribution of 'extrados crack' and 'intrados' cracks overlap significantly with the 'non-crack' distribution, resulting in poor performance. 

These findings highlight the strengths and limitations of each feature type. FPFH is robust for detecting geometric distortions, while the 3D intensity feature improves differentiation between 'non-crack' and 'intrados crack' regions but struggles with 'extrados crack' detection. The 3DMulti-FPFHI leverages the advantages of both, enhancing ‘intrados crack’ detection while mitigating intensity-induced noise, which preserves the 'extrados crack' detection accuracy. In contrast, the 2D feature and CMPF, which rely on image features extracted by the pre-trained neural network, struggle to achieve sufficient performance. Consequently, further analysis will omit 2D intensity feature and CMPF-based detection.

\begin{figure*}[t!]
    \centering
    \includegraphics[width=\linewidth]{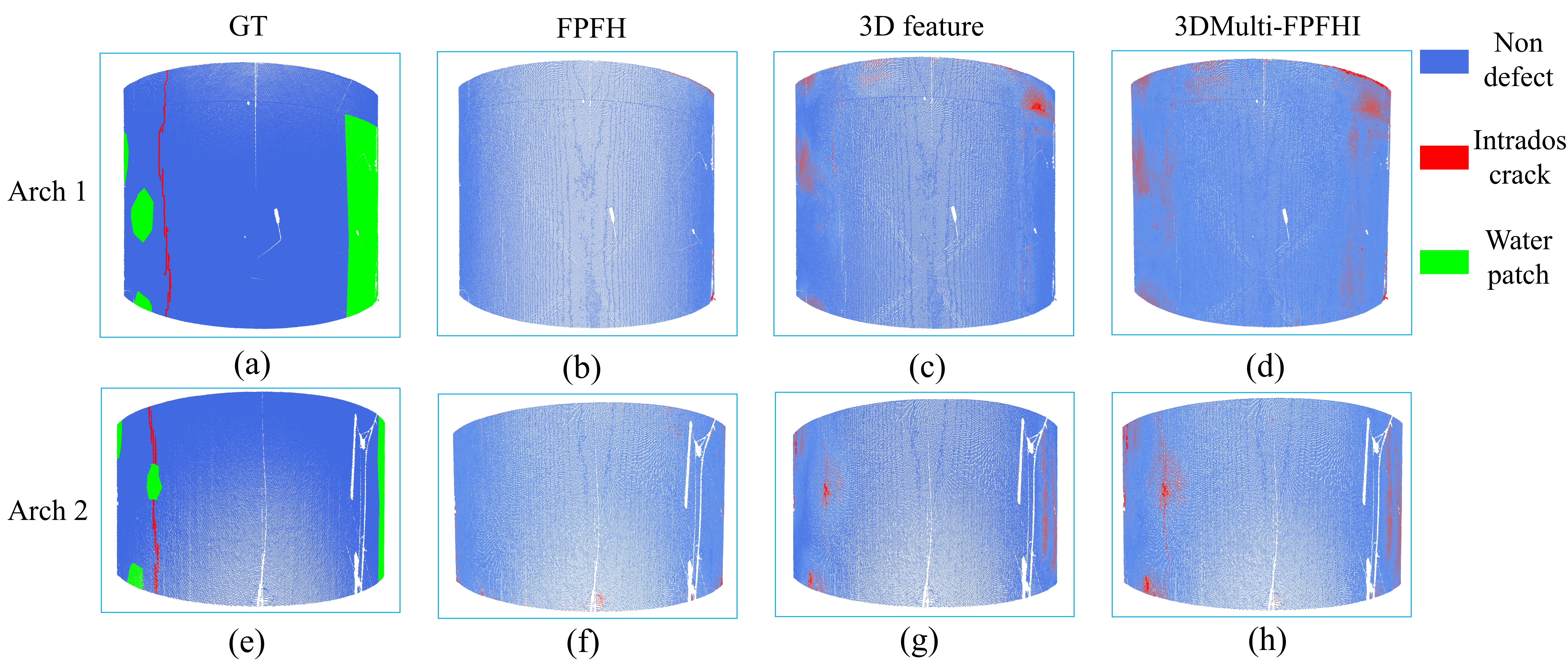}
    \caption{Anomaly detection results for Arch 1 and 2 of the real masonry arch point clouds. (a) and (e) represent the ground truth where blue, red and green colours represent non defect, intrados crack and water patches correspondingly; (b)-(d) and (f)-(h) are anomaly score heatmaps obtained using different feature types, where heatmaps are visualized with the same anomaly scale as in Figure \ref{fig: syn_arch_result_0}.}
    \label{fig: real_arch_result}
\end{figure*}

\begin{table*}[t!]
\centering
\caption{$F1\textit{-}score$ for the real arch}
\label{table: f1 real arch}
\renewcommand{\arraystretch}{1.2}
\begin{tabular} 
 {c|c|c|c}
 \hline
 {} & FPFH & 3D feature & 3DMulti-FPFHI\\
 \hline
 Arch 1$(0.3*s)$ & 0 & \textcolor{blue}{0.029} & \textcolor{red}{0.040} \\
 Arch 1$(0.5*s)$ & 0 & \textcolor{blue}{0.003} & \textcolor{red}{0.011} \\
 \hline
 Arch 2$(0.3*s)$ & 0 & \textcolor{blue}{0.176} & \textcolor{red}{0.289}  \\
 Arch 2$(0.5*s)$ & 0 & \textcolor{blue}{0.052} & \textcolor{red}{0.069}  \\
 \hline
\end{tabular}
\end{table*}

\begin{figure*}[t!]
    \centering
    \includegraphics[width=0.95\linewidth]{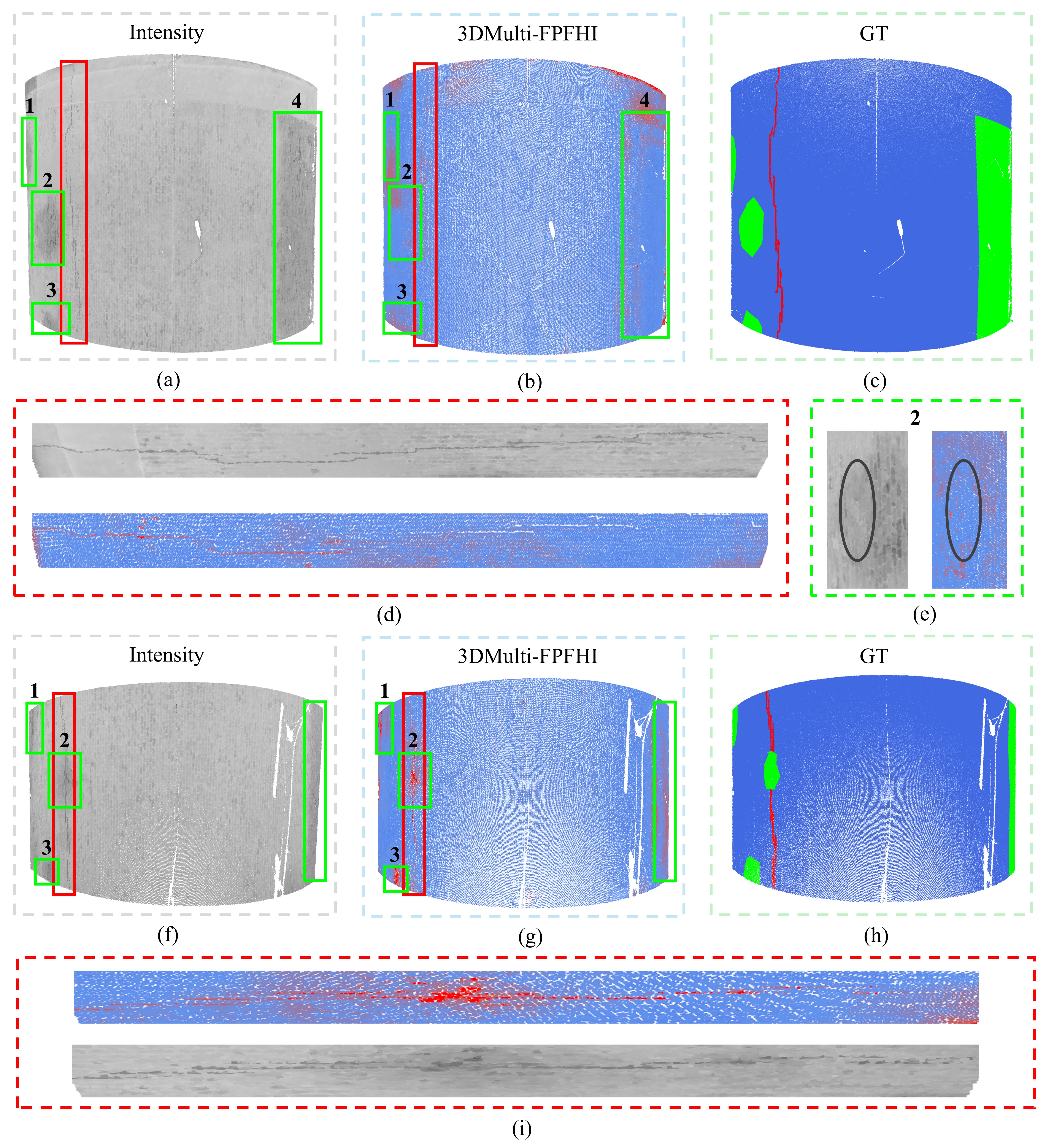}
    \caption{Visualization of 3DMulti-FPFHI in detecting intrados cracks and water patches in Arch 1 and Arch 2. (a)-(c) illustrate the intensity map, 3DMulti-FPFHI detection results, and ground truth for Arch 1, where the intrados crack is highlighted with a red inset, and water patches are labelled from \textbf{1} to \textbf{4} as shown in green insets. (d) provides a close-up view of the intrados crack in Arch 1, shown in both the intensity map and 3DMulti-FPFHI results. (e) show a close-up view of the water patch \textbf{2} in Arch 1, where the black oval indicates the misidentified region; (f)-(h) present the corresponding results for Arch 2, with a detailed view of the intrados crack provided in (i).}
    \label{fig: realarch_zoomed}
\end{figure*}

\subsection{Results for the real arch}

The proposed framework is applied to the real masonry bridge point cloud from London Bridge Station, as shown in Figure \ref{fig: real_arch_result}a-h. Figure \ref{fig: real_arch_result}b and f indicate that FPFH fails to capture the geometric distortion of Arch 1 and Arch 2, resulting in an $F1\textit{-}score$ of 0 (Table \ref{table: f1 real arch}). Previous analysis of this bridge \citep{acikgoz2017evaluation} revealed approximately 2 cm vertical support movement between 05.03.13 and 23.11.13 on one side of Arch E57; the resulting geometric distortion could not be detected by FPFH (consistent with our earlier work \citep{jing4819836anomaly}). Moreover, the intrados crack which appeared was filled with mortar before data collection on 23.11.13, which compromised the detection due to the lack of 'inner crack' points. 

Intrados cracks in Arch 1 and Arch 2 are partially captured using the 3D intensity feature, as shown in Figure \ref{fig: real_arch_result}c and g, respectively. The 3D intensity feature provides additional information for localizing intrados cracks and water patches, as both defect types introduce intensity anomalies into the point clouds. The 3DMulti-FPFHI exhibits consistent detection results with the 3D intensity feature as shown in Figure \ref{fig: real_arch_result}d and h, even though FPFH fails to capture any geometric distortions in Arch 1 and Arch 2. Moreover, the $F1\textit{-}score$ for 3DMulti-FPFHI is higher than that of the 3D intensity feature, as shown in Table \ref{table: f1 real arch}.

Compared to the detection results from the synthetic arch dataset, the 3DMulti-FPFHI shows a much lower $F1\textit{-}score$ in Arch 1 and Arch 2. However, the metric alone cannot fairly evaluate the performance of $PatchCore$ in the real arch dataset due to the diffusion of crack features during detection. It is practically more important to note that the algorithm correctly indicates the locations of intrados cracks and water patches, as further demonstrated in Figure \ref{fig: realarch_zoomed}a-i. The 3DMulti-FPFHI successfully captures intrados cracks in Arch 1 and Arch 2, as highlighted in the red insets in Figure \ref{fig: realarch_zoomed}b and g, where zoomed views of the intensity maps and corresponding detection results accurately delineate intrados cracks. For water patches in Arch 1, the 3DMulti-FPFHI achieves better detection results for smaller patches (e.g., water patches \textbf{1} and \textbf{3}), while providing only vague delineations of the boundaries of larger patches, i.e., water patches \textbf{2} (the zoomed view is provided in Figure \ref{fig: realarch_zoomed}e) and \textbf{4}. A similar trend is observed in Arch 2 (Figure \ref{fig: realarch_zoomed}f-i), where smaller water patches (\textbf{1}-\textbf{3}) are better detected compared to the water patch \textbf{4}.

\begin{figure*}[t!]
    \centering
    \includegraphics[width=\linewidth]{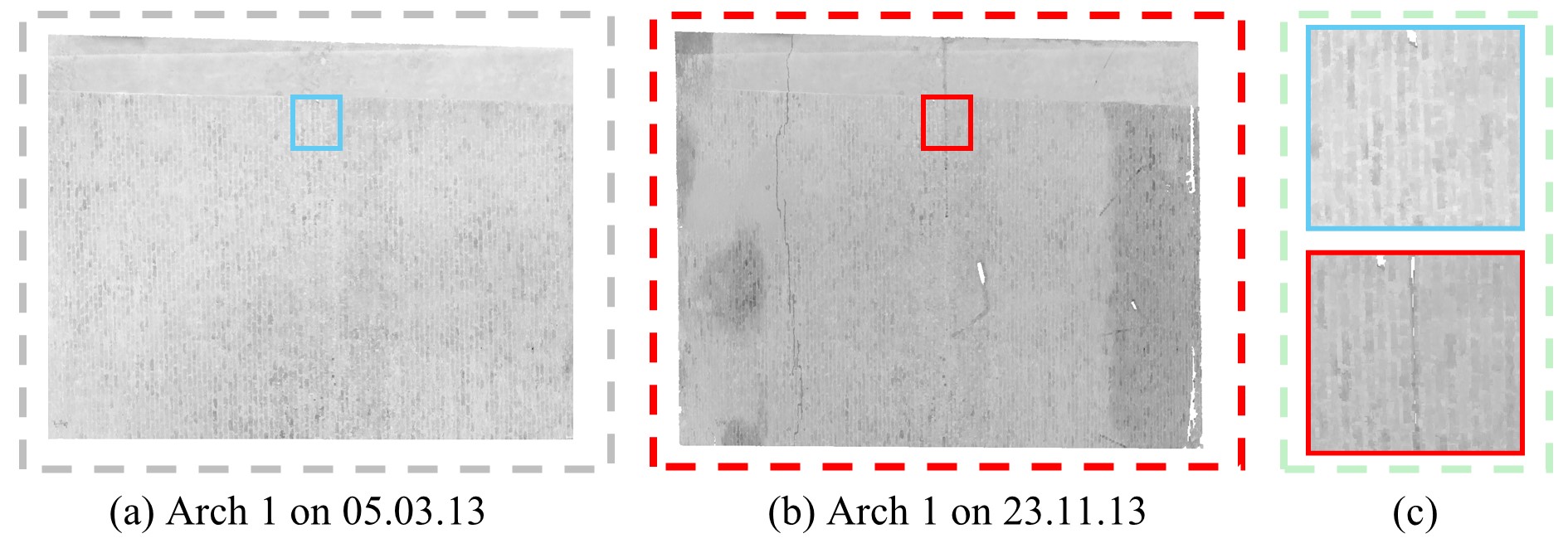}
    \caption{Visualization of intensity map discrepancies in Arch 1. (a) and (b) show the intensity maps of Arch 1 captured on two different dates, with blue and red insets highlighting example regions. (c) illustrates the mismatches in intensity between the two maps, where Arch 2 intensities are represented with darker colours.}
    \label{fig: realarch_mismatching}
\end{figure*} 

The limitations in detecting large water patches can be attributed to the design of the 3D intensity feature, which utilizes absolute relative values between the centre point and its neighbours. This design is to satisfy the underlying assumption of \textit{PatchCore}, namely that $\mathcal{M}_{sub}^*$ can be paired with 3DMulti-FPFHI for 'non defect' points. This assumption holds for synthetic data, where normal and test point clouds are generated under consistent conditions (e.g., surface materials, weather, water leakage, vegetation, etc.), ensuring invariant reflectivity of 'non defect' points across normal and test data. However, in real-world scenarios, these conditions vary significantly over time, impacting the consistency of intensity values for 'non defect' regions. Figure \ref{fig: realarch_mismatching}a-c highlights the intensity mismatches in Arch 1, where the intensity values on 23.11.13 differ significantly from those on 05.03.13 for 'non-crack' points in selected regions.

\begin{figure*}[t!]
    \centering
    \includegraphics[width=\linewidth]{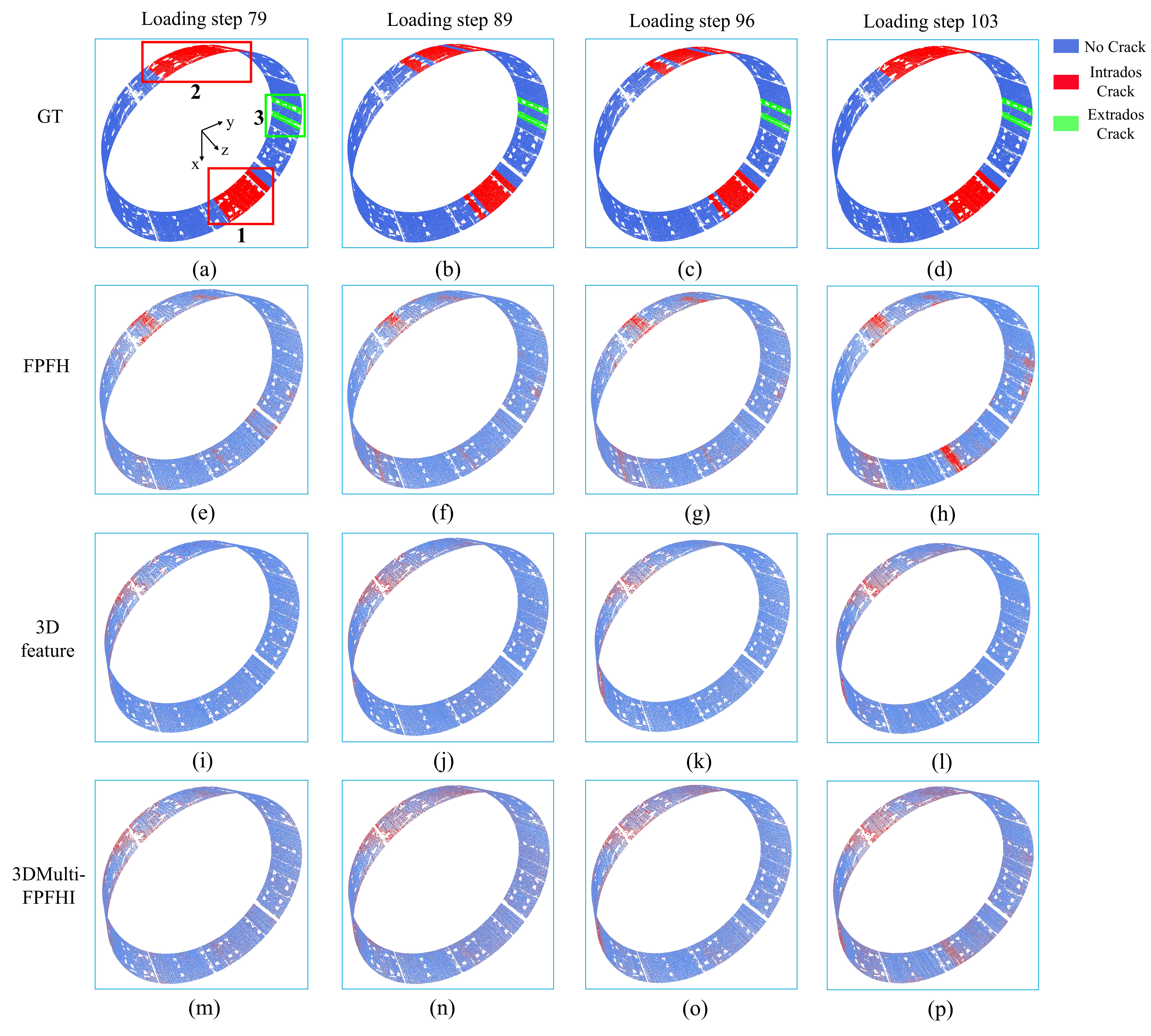}
    \caption{Anomaly detection results corresponding to four different loading steps on the real tunnel lining. (a)-(d) are the ground truth, where intrados and extrados cracks are represented by red and green and indexed \textbf{1}-\textbf{3}, respectively. (e)-(h), (i)-(l), (m)-(p) are heatmaps, where anomaly scale is consistent to Figure \ref{fig: syn_arch_result_0}, representing the anomaly score of each point for the different feature types.}
    \label{fig: real_tunnel}
\end{figure*}

Our 3D intensity feature mitigates the inconsistencies in intensity values caused by environmental effects, by utilizing absolute relative values of intensity to construct the feature. However, this approach becomes less effective when intensity values are uniformly distributed, as the absolute relative differences between the centre point and its neighbours become minimal. Consequently, the 3D intensity feature fails to differentiate between defects and 'non-crack' regions, as demonstrated in the detection of water patch \textbf{2} in Arch 1, highlighted by the black oval in Figure \ref{fig: realarch_zoomed}e. This limitation explains why the 3D intensity feature primarily detects defects along the boundaries of large water patches, where the intensity gradient is highest, while failing to identify anomalies within the patch interior.

Despite this limitation, it is important to emphasize that incorporating the 3D intensity feature enables the detection of intrados cracks and water patches—an achievement not possible using geometric features alone. Future studies could address this limitation by employing more sensitive and robust 3D feature extraction techniques.

\begin{table*}[t!]
\centering
\caption{$F1\textit{-}score$ for the real tunnel}
\label{table: f1 real tunnel}
\renewcommand{\arraystretch}{1.2}
\begin{tabular} 
 {M{4.5cm}|M{1.5cm}|M{3cm}|M{3cm}}
 \hline
 {} & FPFH & 3D feature & 3DMulti-FPFHI \\
 \hline
 Loading step 79$(0.3*s)$ & \textcolor{red}{0.057} & 0.025 & \textcolor{blue}{0.025} \\
 Loading step 79$(0.5*s)$ & 0.001 & \textcolor{blue}{0.006} & \textcolor{red}{0.006} \\
 \hline
 Loading step 89$(0.3*s)$ & 0.037 & \textcolor{blue}{0.044} & \textcolor{red}{0.162} \\
 Loading step 89$(0.5*s)$ & 0.001 & 0.011 & 0.022 \\
 \hline
 Loading step 96$(0.3*s)$ & \textcolor{blue}{0.088} & 0.030 & \textcolor{red}{0.093} \\
 Loading step 96$(0.5*s)$ & \textcolor{blue}{0.009} & 0.007 & \textcolor{red}{0.011} \\
 \hline
 Loading step 103$(0.3*s)$ & \textcolor{blue}{0.116} & 0.052 & \textcolor{red}{0.206} \\
 Loading step 103$(0.5*s)$ & \textcolor{red}{0.043} & 0.015 & \textcolor{blue}{0.026} \\
 \hline
\end{tabular}
\end{table*}

\subsection{Results for the experimental tunnel model}

The proposed framework is validated using the point clouds of an experimental tunnel model. Point clouds from the loading steps 79, 89, 96, and 103 are used; results are shown in Figure \ref{fig: real_tunnel}a-d and e-p, respectively. The undeformed point cloud at loading step 0 is used as the 'normal' data.

Intrados and extrados cracks are visualized and indexed as \textbf{1}-\textbf{3} in Figure \ref{fig: real_tunnel}a-d (consistent with Figure \ref{fig: bird_tunnel}b). 

Anomaly detection results using FPFH are shown in Figure \ref{fig: real_tunnel}e-h, where cracks are identified using only geometric distortions. FPFH shows low $F1\textit{-}score$ across all loading steps (Table \ref{table: f1 real tunnel}). As in Section 4.2, the $F1\textit{-}score$ inadequately reflects the performance of \textit{PatchCore}. This is partially caused by the diffusion of the predicted crack regions and the mesh-based label mapping of the ground truth.

\begin{figure*}[t!]
    \centering
    \includegraphics[width=\linewidth]{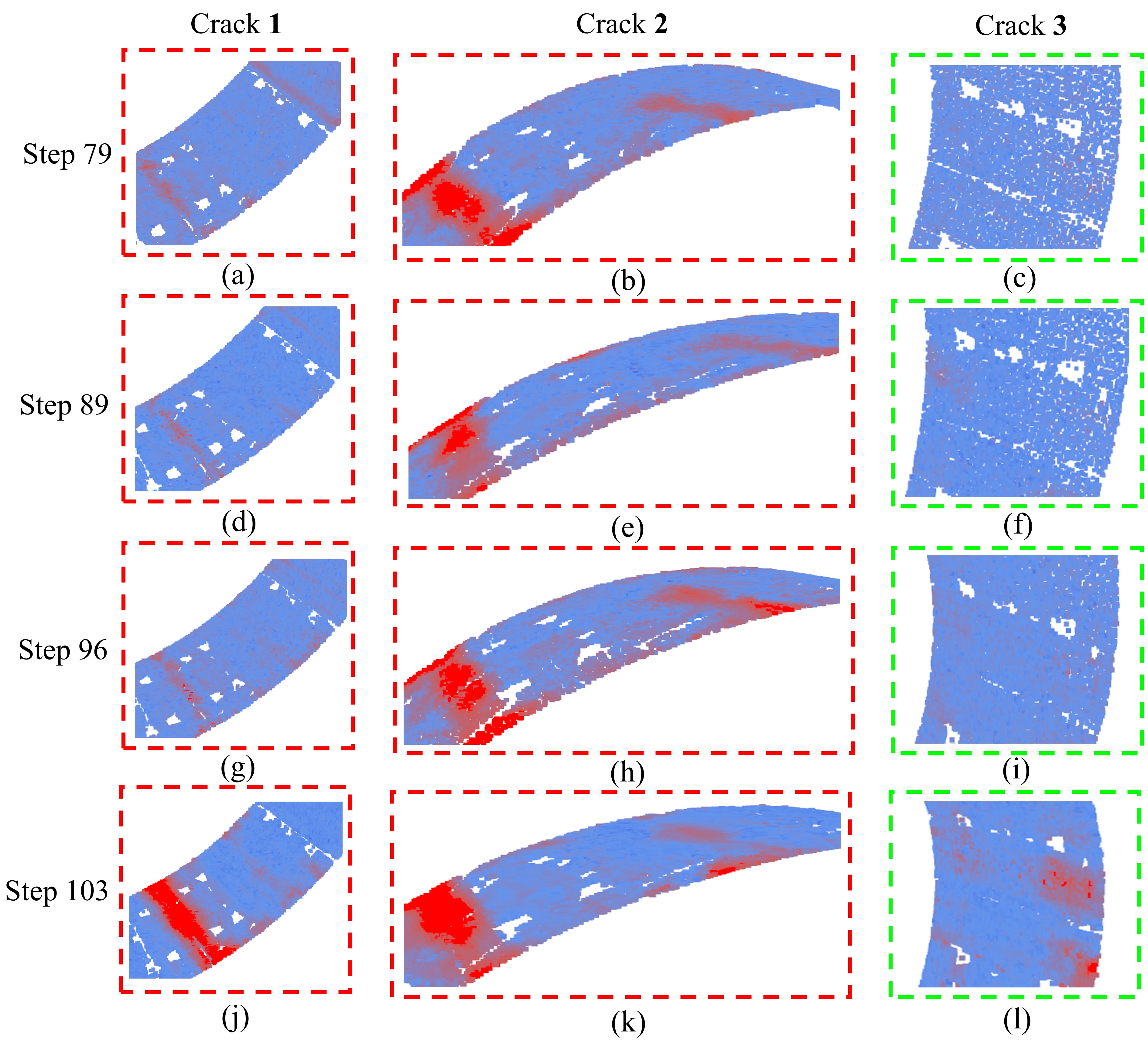}
    \caption{Visualization of zoomed crack detection results using FPFH. (a)-(c), (d)-(f), (g)-(i), and (j)-(l) present zoomed views of cracks \textbf{1}-\textbf{3} (as defined in Figure \ref{fig: real_tunnel}a) at loading steps 79, 89, 96, and 103, respectively.}
    \label{fig: realtunnel_cracks_zoomed}
\end{figure*}

To demonstrate the feasibility of \textit{PatchCore}, zoomed views of cracks \textbf{1}-\textbf{3} are provided for FPFH detection results at different loading steps, as shown in Figure \ref{fig: realtunnel_cracks_zoomed}a-l. Regions corresponding to two intrados cracks \textbf{1} and \textbf{2} can be localized by leveraging only geometric information as demonstrated in Figure \ref{fig: realtunnel_cracks_zoomed}a, d, g, j, b, e, h, and k. The anomaly scales at loading steps 79, 89, and 96 show no significant improvements due to small variations of geometry during crack development. However, a noticeable improvement in anomaly scales is observed at loading step 103, where severe damage is introduced.

\begin{figure*}[t!]
    \centering
    \includegraphics[width=\linewidth]{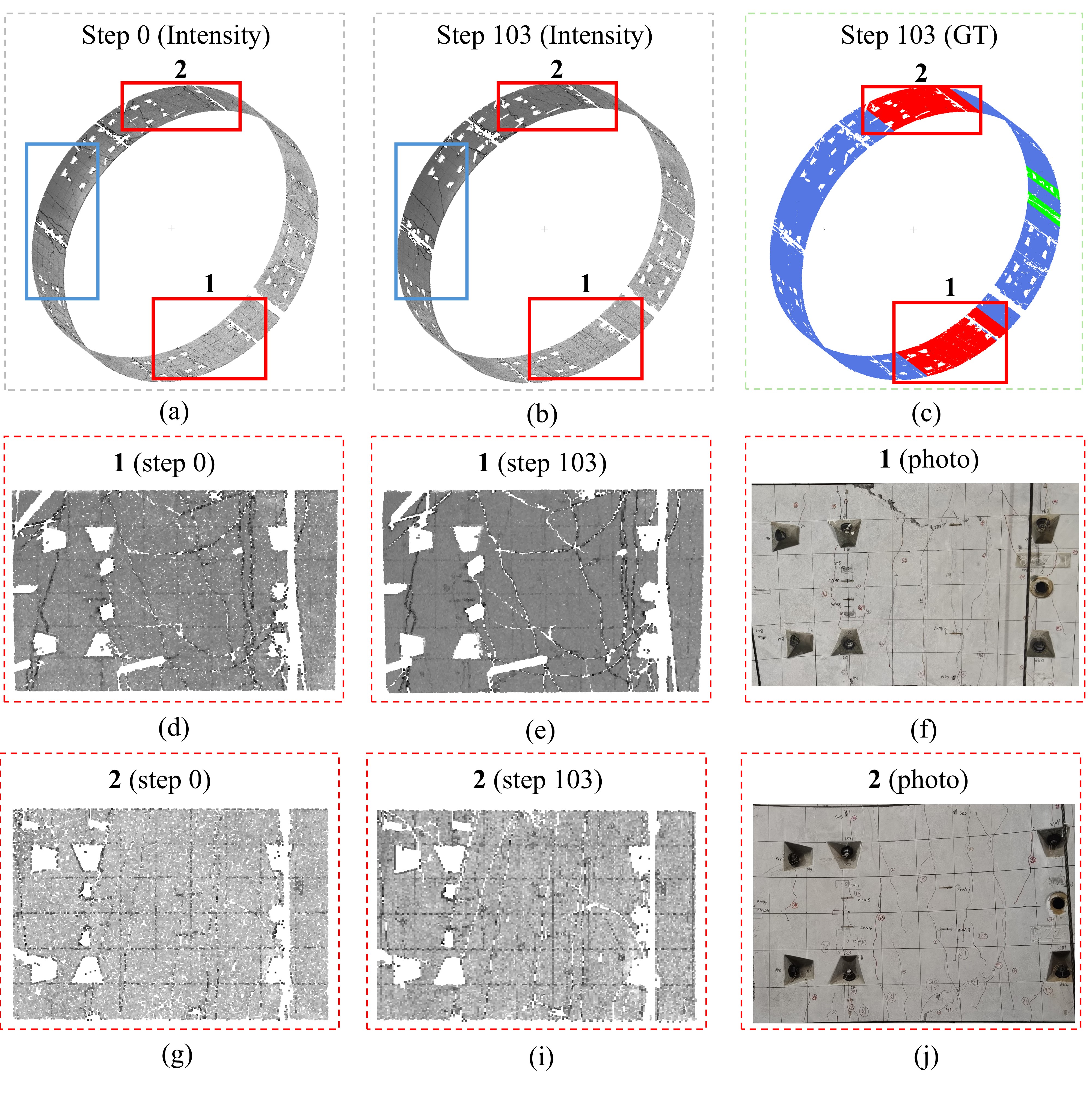}
    \caption{Limitations of intensity information in detecting cracks in the real tunnel point cloud. (a) and (b) represent the tunnel in loading steps 0 and 103 by transferring intensity information into the greyscale. Blue boxes indicate inconsistency of the intensity between step 0 and 103; (c) shows the ground truth of cracks \textbf{1} and \textbf{2}; (d) and (e) show insets for crack \textbf{1} for step 0 and 103; (f) shows the photo of crack \textbf{1}; and (g)-(j) shows the corresponding insets and photo for crack \textbf{2}.}
    \label{fig: real_tunnel_failed}
\end{figure*}

The extrados crack \textbf{3} initially shows faint signals at loading step 89 and gradually becomes more visible by step 103 (as shown in Figure \ref{fig: realtunnel_cracks_zoomed}c, f, i, and l). Loading step 103 highlights geometric distortions introduced by the crack \textbf{3}. Extrados tunnel cracks are often missed in defect detection tasks (e.g., using image-based algorithms) due to insufficient surface features. In contrast, FPFH provides a potential method for identifying extrados cracks by examining the geometric information indirectly.

In contrast, the 3D intensity feature fails to provide reliable detection results for the experimental tunnel dataset, resulting in the lowest $F1\textit{-}score$ in most cases as shown in Table \ref{table: f1 real tunnel}. To investigate the cause, intensity maps (greyscale) at loading steps 0 and 103 are visualized in Figure \ref{fig: real_tunnel_failed}a-b. Ground truth in Figure \ref{fig: real_tunnel_failed}c includes cracks \textbf{1} and \textbf{2}, which are visible to the laser scanner. Insets in Figure \ref{fig: real_tunnel_failed}d and e show zoomed views of regions corresponding to crack \textbf{1} at steps 0 and 103. The photo of crack \textbf{1} at step 103 is provided in Figure \ref{fig: real_tunnel_failed}f. 

\begin{figure*}[t!]
    \centering
    \includegraphics[width=\linewidth]{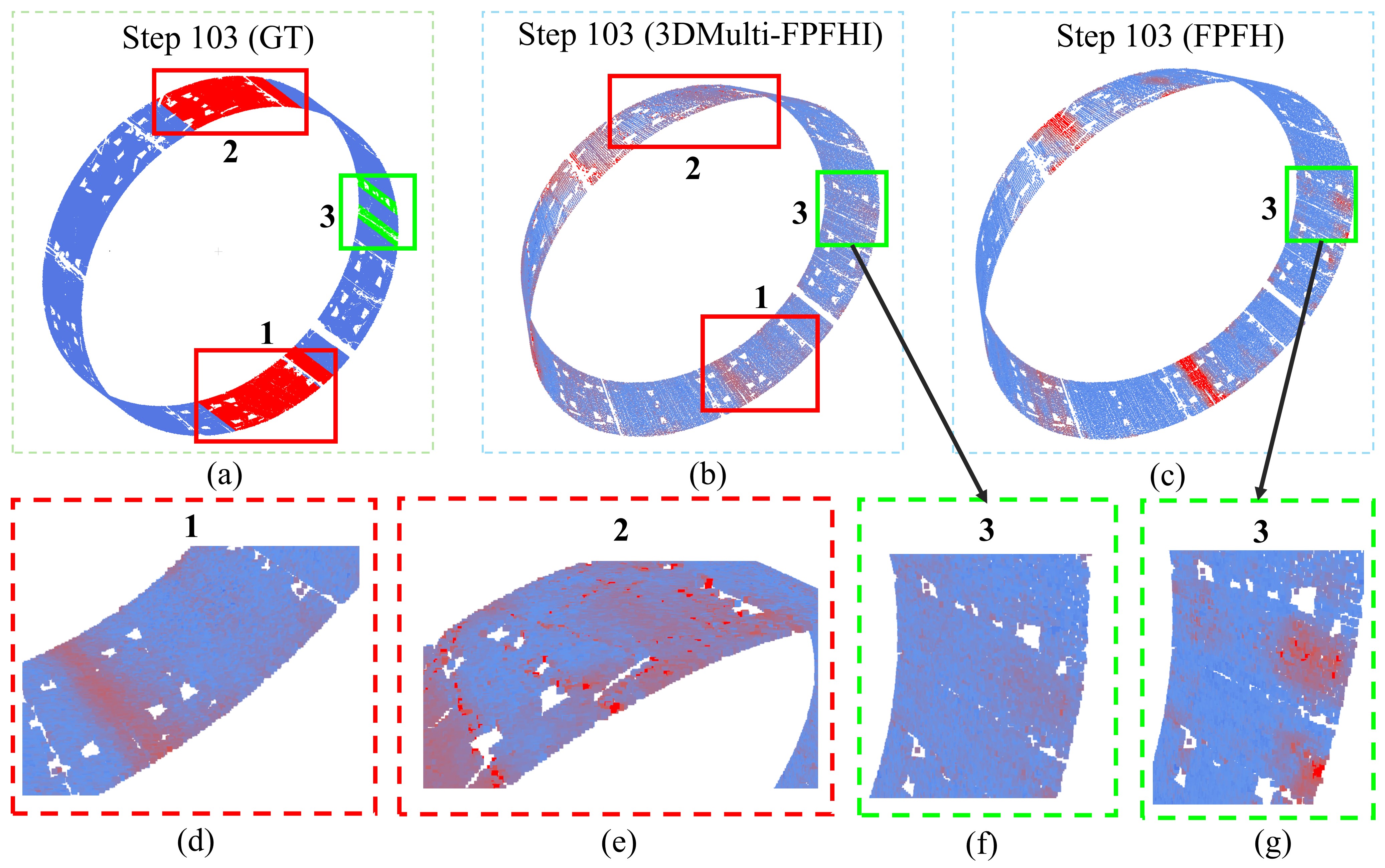}
    \caption{Comparison of detection results using FPFH and 3DMulti-FPFHI at loading step 103. (a), (b), and (c) depict the ground truth, FPFH detection result, and 3DMulti-FPFHI detection result, respectively. (d), (e), and (f) present zoomed views of cracks \textbf{1}-\textbf{3} for 3DMulti-FPFHI, while (g) shows the zoomed view of crack \textbf{3} for FPFH.}
    \label{fig: realtunnel_cracks_feature_compare}
\end{figure*}

By comparing Figure \ref{fig: real_tunnel_failed}d and e, no distinct intensity differentiations can be observed that indicate the presence of cracks or any different defect types seen in Figure \ref{fig: real_tunnel_failed}f. The visible black lines represent cables, not cracks, suggesting that the intensity information failed to distinguish cracks from nearby normal regions, as seen in real arch dataset results. The primary reason for this limitation is that the scanning resolution is lower than the size of the cracks, resulting in edge effects \citep{stalowska2022crack}. Similar edge effects are also evident in crack \textbf{2}, as shown in Figure \ref{fig: real_tunnel_failed}g-j. Meanwhile, the 3D intensity feature introduces detection noise into regions indicated by blue boxes in Figure \ref{fig: real_tunnel_failed}a-b.

Results corresponding to 3DMulti-FPFHI show a similar tendency as FPFH, as 3DMulti-FPFHI can mitigate intensity inconsistencies introduced by the 3D intensity feature (Section 4.3). To demonstrate this, Figure \ref{fig: realtunnel_cracks_feature_compare}a-g compares FPFH and 3DMulti-FPFHI detection results at loading step 103. The detection of intrados cracks \textbf{1} and \textbf{2} by the 3DMulti-FPFHI are only slightly compromised compared to FPFH. However, the fusion of the 3D intensity feature introduces noise and diffuses defects (Figure \ref{fig: real_arch_result}i-l). The introduction of the 3D intensity feature compromises FPFH performance in the extrados crack \textbf{3} as shown in Figure \ref{fig: realtunnel_cracks_feature_compare}f and g, despite the $F1\textit{-}score$ (see Table \ref{table: f1 real tunnel}) being preserved. Nonetheless, the 3DMulti-FPFHI demonstrates reasonable performance across all loading steps in capturing intrados cracks, maintaining a degree of robustness despite intensity information introducing significant noise.

In an ideal case, $min\_vals$ for 3D intensity feature, calculated by pairing $\boldsymbol{L}_{3DI}$ and 3D intensity feature in $\mathcal{M}_{sub}^*$, should be approximately zero if not detecting defects. For example, as demonstrated in Figure \ref{fig: real_arch_result}b and h, FPFH failed to detect any anomalies in the real masonry arch, which consequently did not degrade the performance of the 3DMulti-FPHFI.

\section*{Conclusions}

This paper presents a novel 3D multimodal feature, 3DMulti-FPFHI, which fuses two modalities, i.e., geometry and intensity information from point clouds, for use with the \textit{PatchCore} anomaly detection algorithm to identify defects from different types of infrastructure assets. To effectively leverage intensity information, we propose a 3D intensity feature that embeds intensity values by computing the absolute relative differences between the centre point and its neighbours. The developed framework is then applied to detect defects in synthetic and real point cloud datasets. The proposed method was benchmarked against standard FPFH and the SOTA multimodal anomaly detection method CPMF \citep{cao2023complementary}, which uses a pre-trained ResNet50d to extract 2D features from point cloud projections. The influence of the 3D intensity feature was also analyzed independently to assess the contribution of this modality to anomaly detection. A key advantage of the proposed framework over learning-based methods is that the method does not rely on extensive training to identify defects. Instead, it only requires two pairs of point clouds.

We first modified a previous synthetic masonry arch dataset by incorporating surface roughness and the geometry of intrados cracks to better reflect real-world conditions. The performance of the anomaly detection algorithm was re-evaluated using various feature types, demonstrating that the inclusion of 3D intensity information significantly improves the detection of intrados cracks. Performance discrepancies among different feature types were also analysed using the distribution of anomaly scores (ranging from 0 to 1, representing a low-to-high anomaly scale) and comparing their detection accuracy for intrados and extrados cracks. The analysis demonstrated how intensity-related anomalies assist detection quantitatively and explained why the 2D intensity feature and CMPF fail to provide accurate results.

The framework was further validated on real masonry bridge point cloud data from London Bridge Station, collected on two different dates (05.03.13 and 23.11.13). An intrados crack was identified in the 23.11.13 dataset after maintenance work, with 3DMulti-FPFHI demonstrating accurate detection results. By integrating the 3D intensity feature, 3DMulti-FPFHI successfully captured water patches that introduced intensity anomalies, expanding the ability of $PatchCore$ to detect non-geometric defects. Furthermore, the 3D intensity feature effectively mitigated inconsistencies in regions without defects caused by environmental variations between the two datasets, ensuring reasonable localization of both cracks and water patches.

To evaluate the generalization capability of 3DMulti-FPFHI, we applied the framework to a full-scale experimental tunnel dataset subjected to increasing loading steps. Using an undeformed point cloud (loading step 0) as the reference, the algorithm successfully identified intrados cracks. FPFH also detected extrados cracks, which are often overlooked due to a lack of associated surface features. While FPFH achieved better results in this case as the 3D intensity feature failed to provide useful information, 3DMulti-FPFHI still provided satisfactory performance despite the intensity noise introduced by the experimental equipment.

Through evaluations across different datasets, we demonstrate that 3DMulti-FPFHI offers a robust solution by aggregating geometric and intensity modalities, particularly in real-world scenarios with environmental inconsistencies (e.g., due to atmospheric changes in air temperature, relative humidity, sunlight) which challenge the underlying assumptions of anomaly detection methods. Additionally, its potential for generalization to various types of infrastructure has been demonstrated. 


\section*{Limitations}

The 3D multimodal feature failed to capture small cracks in the synthetic dataset (e.g., intrados and extrados cracks with maximum values of 2mm and 6mm respectively) and the real tunnel (the overall range of crack width is from 0.02 mm to several millimetres). From a geometric perspective, small cracks cause subtle distortions that FPFH cannot capture. Additionally, the intrinsic sparsity of point clouds limits their resolution compared to 2D images, making it harder to identify fine details. Therefore, the 3D intensity feature is affected not only by noise but also by insufficient resolution, reducing their ability to detect small cracks accurately.

Meanwhile, the current anomaly detection framework fails to provide an optimal boundary for distinguishing 'non-defect' and 'defect' points. Unsupervised learning methods such as DBSCAN or K-clustering require frequent hyperparameter calibration due to variations in anomaly score distributions across datasets, limiting their generalizability.

Future work could focus on developing more sensitive geometric indicators to detect subtle deformations, potentially leveraging learning-based approaches (e.g., \citep{deng2018ppfnet}). Additionally, incorporating methods like one-class support vector machines into \textit{PatchCore} could help derive optimal detection boundaries and enhance anomaly detection performance.

\section*{Acknowledgement}

The authors acknowledge Fan Zhang (Tongji University) for sharing the image and crack data for the real tunnel.

\bibliographystyle{elsarticle-harv}     
\bibliography{update_patchcore_citations}

\begin{thebibliography}{59}
\expandafter\ifx\csname natexlab\endcsname\relax\def\natexlab#1{#1}\fi
\providecommand{\url}[1]{\texttt{#1}}
\providecommand{\href}[2]{#2}
\providecommand{\path}[1]{#1}
\providecommand{\DOIprefix}{doi:}
\providecommand{\ArXivprefix}{arXiv:}
\providecommand{\URLprefix}{URL: }
\providecommand{\Pubmedprefix}{pmid:}
\providecommand{\doi}[1]{\href{http://dx.doi.org/#1}{\path{#1}}}
\providecommand{\Pubmed}[1]{\href{pmid:#1}{\path{#1}}}
\providecommand{\bibinfo}[2]{#2}
\ifx\xfnm\relax \def\xfnm[#1]{\unskip,\space#1}\fi
\bibitem[{Acikgoz et~al.(2018)Acikgoz, DeJong and Soga}]{acikgoz2018sensing}
\bibinfo{author}{Acikgoz, S.}, \bibinfo{author}{DeJong, M.J.}, \bibinfo{author}{Soga, K.}, \bibinfo{year}{2018}.
\newblock \bibinfo{title}{Sensing dynamic displacements in masonry rail bridges using 2d digital image correlation}.
\newblock \bibinfo{journal}{Structural Control and Health Monitoring} \bibinfo{volume}{25}, \bibinfo{pages}{e2187}.
\newblock \URLprefix \url{https://doi.org/10.1002/stc.2187}.
\bibitem[{Acikgoz et~al.(2017)Acikgoz, Soga and Woodhams}]{acikgoz2017evaluation}
\bibinfo{author}{Acikgoz, S.}, \bibinfo{author}{Soga, K.}, \bibinfo{author}{Woodhams, J.}, \bibinfo{year}{2017}.
\newblock \bibinfo{title}{Evaluation of the response of a vaulted masonry structure to differential settlements using point cloud data and limit analyses}.
\newblock \bibinfo{journal}{Construction and Building Materials} \bibinfo{volume}{150}, \bibinfo{pages}{916--931}.
\newblock \URLprefix \url{https://doi.org/10.1016/j.conbuildmat.2017.05.075}.
\bibitem[{Akcay et~al.(2019)Akcay, Atapour-Abarghouei and Breckon}]{akcay2019ganomaly}
\bibinfo{author}{Akcay, S.}, \bibinfo{author}{Atapour-Abarghouei, A.}, \bibinfo{author}{Breckon, T.P.}, \bibinfo{year}{2019}.
\newblock \bibinfo{title}{Ganomaly: Semi-supervised anomaly detection via adversarial training}, in: \bibinfo{booktitle}{Computer Vision--ACCV 2018: 14th Asian Conference on Computer Vision, Perth, Australia, December 2--6, 2018, Revised Selected Papers, Part III 14}, \bibinfo{organization}{Springer}. pp. \bibinfo{pages}{622--637}.
\newblock \URLprefix \url{https://doi.org/10.1016/j.media.2019.01.010}.
\bibitem[{Attard et~al.(2019)Attard, Debono, Valentino, Di~Castro, Masi and Scibile}]{attard2019automatic}
\bibinfo{author}{Attard, L.}, \bibinfo{author}{Debono, C.J.}, \bibinfo{author}{Valentino, G.}, \bibinfo{author}{Di~Castro, M.}, \bibinfo{author}{Masi, A.}, \bibinfo{author}{Scibile, L.}, \bibinfo{year}{2019}.
\newblock \bibinfo{title}{Automatic crack detection using mask r-cnn}, in: \bibinfo{booktitle}{2019 11th international symposium on image and signal processing and analysis (ISPA)}, \bibinfo{organization}{IEEE}. pp. \bibinfo{pages}{152--157}.
\newblock \URLprefix \url{10.1109/ISPA.2019.8868619}.
\bibitem[{Beggel et~al.(2020)Beggel, Pfeiffer and Bischl}]{beggel2020robust}
\bibinfo{author}{Beggel, L.}, \bibinfo{author}{Pfeiffer, M.}, \bibinfo{author}{Bischl, B.}, \bibinfo{year}{2020}.
\newblock \bibinfo{title}{Robust anomaly detection in images using adversarial autoencoders}, in: \bibinfo{booktitle}{Machine Learning and Knowledge Discovery in Databases: European Conference, ECML PKDD 2019, W{\"u}rzburg, Germany, September 16--20, 2019, Proceedings, Part I}, \bibinfo{organization}{Springer}. pp. \bibinfo{pages}{206--222}.
\newblock \URLprefix \url{https://doi.org/10.1007/978-3-030-46150-8_13}.
\bibitem[{Bergmann et~al.(2021)Bergmann, Jin, Sattlegger and Steger}]{bergmann2021mvtec}
\bibinfo{author}{Bergmann, P.}, \bibinfo{author}{Jin, X.}, \bibinfo{author}{Sattlegger, D.}, \bibinfo{author}{Steger, C.}, \bibinfo{year}{2021}.
\newblock \bibinfo{title}{The mvtec 3d-ad dataset for unsupervised 3d anomaly detection and localization}.
\newblock \bibinfo{journal}{arXiv preprint arXiv:2112.09045} \URLprefix \url{https://doi.org/10.5220/0010865000003124}.
\bibitem[{Brackenbury(2022)}]{BrackenburyDaniel2022AIIo}
\bibinfo{author}{Brackenbury, D.}, \bibinfo{year}{2022}.
\newblock \bibinfo{title}{Automated image-based inspection of masonry arch bridges}.
\bibitem[{Cao et~al.(2022)Cao, Wan, Shen and Gao}]{cao2022informative}
\bibinfo{author}{Cao, Y.}, \bibinfo{author}{Wan, Q.}, \bibinfo{author}{Shen, W.}, \bibinfo{author}{Gao, L.}, \bibinfo{year}{2022}.
\newblock \bibinfo{title}{Informative knowledge distillation for image anomaly segmentation}.
\newblock \bibinfo{journal}{Knowledge-Based Systems} \bibinfo{volume}{248}, \bibinfo{pages}{108846}.
\newblock \URLprefix \url{https://doi.org/10.1016/j.knosys.2022.108846}.
\bibitem[{Cao et~al.(2023)Cao, Xu and Shen}]{cao2023complementary}
\bibinfo{author}{Cao, Y.}, \bibinfo{author}{Xu, X.}, \bibinfo{author}{Shen, W.}, \bibinfo{year}{2023}.
\newblock \bibinfo{title}{Complementary pseudo multimodal feature for point cloud anomaly detection}.
\newblock \bibinfo{journal}{arXiv preprint arXiv:2303.13194} \URLprefix \url{https://doi.org/10.48550/arXiv.2303.13194}.
\bibitem[{Cha et~al.(2017)Cha, Choi and B{\"u}y{\"u}k{\"o}zt{\"u}rk}]{cha2017deep}
\bibinfo{author}{Cha, Y.J.}, \bibinfo{author}{Choi, W.}, \bibinfo{author}{B{\"u}y{\"u}k{\"o}zt{\"u}rk, O.}, \bibinfo{year}{2017}.
\newblock \bibinfo{title}{Deep learning-based crack damage detection using convolutional neural networks}.
\newblock \bibinfo{journal}{Computer-Aided Civil and Infrastructure Engineering} \bibinfo{volume}{32}, \bibinfo{pages}{361--378}.
\newblock \URLprefix \url{https://doi.org/10.1111/mice.12263}.
\bibitem[{Cha et~al.(2018)Cha, Choi, Suh, Mahmoudkhani and B{\"u}y{\"u}k{\"o}zt{\"u}rk}]{cha2018autonomous}
\bibinfo{author}{Cha, Y.J.}, \bibinfo{author}{Choi, W.}, \bibinfo{author}{Suh, G.}, \bibinfo{author}{Mahmoudkhani, S.}, \bibinfo{author}{B{\"u}y{\"u}k{\"o}zt{\"u}rk, O.}, \bibinfo{year}{2018}.
\newblock \bibinfo{title}{Autonomous structural visual inspection using region-based deep learning for detecting multiple damage types}.
\newblock \bibinfo{journal}{Computer-Aided Civil and Infrastructure Engineering} \bibinfo{volume}{33}, \bibinfo{pages}{731--747}.
\newblock \URLprefix \url{https://doi.org/10.1111/mice.12334}.
\bibitem[{Chen and Jahanshahi(2017)}]{chen2017nb}
\bibinfo{author}{Chen, F.C.}, \bibinfo{author}{Jahanshahi, M.R.}, \bibinfo{year}{2017}.
\newblock \bibinfo{title}{Nb-cnn: Deep learning-based crack detection using convolutional neural network and na{\"\i}ve bayes data fusion}.
\newblock \bibinfo{journal}{IEEE Transactions on Industrial Electronics} \bibinfo{volume}{65}, \bibinfo{pages}{4392--4400}.
\newblock \URLprefix \url{10.1109/TIE.2017.2764844}.
\bibitem[{Costanzino et~al.(2024)Costanzino, Ramirez, Lisanti and Di~Stefano}]{costanzino2024multimodal}
\bibinfo{author}{Costanzino, A.}, \bibinfo{author}{Ramirez, P.Z.}, \bibinfo{author}{Lisanti, G.}, \bibinfo{author}{Di~Stefano, L.}, \bibinfo{year}{2024}.
\newblock \bibinfo{title}{Multimodal industrial anomaly detection by crossmodal feature mapping}, in: \bibinfo{booktitle}{Proceedings of the IEEE/CVF Conference on Computer Vision and Pattern Recognition}, pp. \bibinfo{pages}{17234--17243}.
\newblock \URLprefix \url{https://doi.org/10.48550/arXiv.2312.04521}.
\bibitem[{Dais et~al.(2021)Dais, Bal, Smyrou and Sarhosis}]{dais2021automatic}
\bibinfo{author}{Dais, D.}, \bibinfo{author}{Bal, I.E.}, \bibinfo{author}{Smyrou, E.}, \bibinfo{author}{Sarhosis, V.}, \bibinfo{year}{2021}.
\newblock \bibinfo{title}{Automatic crack claslogoglu2016cospairsification and segmentation on masonry surfaces using convolutional neural networks and transfer learning}.
\newblock \bibinfo{journal}{Automation in Construction} \bibinfo{volume}{125}, \bibinfo{pages}{103606}.
\newblock \URLprefix \url{https://doi.org/10.1016/j.autcon.2021.103606}.
\bibitem[{Deng et~al.(2018)Deng, Birdal and Ilic}]{deng2018ppfnet}
\bibinfo{author}{Deng, H.}, \bibinfo{author}{Birdal, T.}, \bibinfo{author}{Ilic, S.}, \bibinfo{year}{2018}.
\newblock \bibinfo{title}{Ppfnet: Global context aware local features for robust 3d point matching}, in: \bibinfo{booktitle}{Proceedings of the IEEE conference on computer vision and pattern recognition}, pp. \bibinfo{pages}{195--205}.
\newblock \URLprefix \url{https://doi.org/10.48550/arXiv.1802.02669}.
\bibitem[{Deng et~al.(2009)Deng, Dong, Socher, Li, Li and Fei-Fei}]{5206848}
\bibinfo{author}{Deng, J.}, \bibinfo{author}{Dong, W.}, \bibinfo{author}{Socher, R.}, \bibinfo{author}{Li, L.J.}, \bibinfo{author}{Li, K.}, \bibinfo{author}{Fei-Fei, L.}, \bibinfo{year}{2009}.
\newblock \bibinfo{title}{Imagenet: A large-scale hierarchical image database}, in: \bibinfo{booktitle}{2009 IEEE Conference on Computer Vision and Pattern Recognition}, pp. \bibinfo{pages}{248--255}.
\newblock \DOIprefix\doi{10.1109/CVPR.2009.5206848}.
\bibitem[{Dong et~al.(2023)Dong, Wang, Chen, Jiang, Li and Gu}]{dong2023pavement}
\bibinfo{author}{Dong, Q.}, \bibinfo{author}{Wang, S.}, \bibinfo{author}{Chen, X.}, \bibinfo{author}{Jiang, W.}, \bibinfo{author}{Li, R.}, \bibinfo{author}{Gu, X.}, \bibinfo{year}{2023}.
\newblock \bibinfo{title}{Pavement crack detection based on point cloud data and data fusion}.
\newblock \bibinfo{journal}{Philosophical Transactions of the Royal Society A} \bibinfo{volume}{381}, \bibinfo{pages}{20220165}.
\newblock \URLprefix \url{https://doi.org/10.1098/rsta.2022.0165}.
\bibitem[{Fan et~al.(2020)Fan, Li, Chen, Wei, Loprencipe, Chen and Di~Mascio}]{fan2020automatic}
\bibinfo{author}{Fan, Z.}, \bibinfo{author}{Li, C.}, \bibinfo{author}{Chen, Y.}, \bibinfo{author}{Wei, J.}, \bibinfo{author}{Loprencipe, G.}, \bibinfo{author}{Chen, X.}, \bibinfo{author}{Di~Mascio, P.}, \bibinfo{year}{2020}.
\newblock \bibinfo{title}{Automatic crack detection on road pavements using encoder-decoder architecture}.
\newblock \bibinfo{journal}{Materials} \bibinfo{volume}{13}, \bibinfo{pages}{2960}.
\newblock \URLprefix \url{https://doi.org/10.3390/ma13132960}.
\bibitem[{Gupta and Dixit(2022)}]{gupta2022image}
\bibinfo{author}{Gupta, P.}, \bibinfo{author}{Dixit, M.}, \bibinfo{year}{2022}.
\newblock \bibinfo{title}{Image-based crack detection approaches: a comprehensive survey}.
\newblock \bibinfo{journal}{Multimedia Tools and Applications} \bibinfo{volume}{81}, \bibinfo{pages}{40181--40229}.
\newblock \URLprefix \url{https://doi.org/10.1007/s11042-022-13152-z}.
\bibitem[{Hallee et~al.(2021)Hallee, Napolitano, Reinhart and Glisic}]{hallee2021crack}
\bibinfo{author}{Hallee, M.J.}, \bibinfo{author}{Napolitano, R.K.}, \bibinfo{author}{Reinhart, W.F.}, \bibinfo{author}{Glisic, B.}, \bibinfo{year}{2021}.
\newblock \bibinfo{title}{Crack detection in images of masonry using cnns}.
\newblock \bibinfo{journal}{Sensors} \bibinfo{volume}{21}, \bibinfo{pages}{4929}.
\newblock \URLprefix \url{https://doi.org/10.3390/s21144929}.
\bibitem[{He et~al.(2016)He, Zhang, Ren and Sun}]{he2016deep}
\bibinfo{author}{He, K.}, \bibinfo{author}{Zhang, X.}, \bibinfo{author}{Ren, S.}, \bibinfo{author}{Sun, J.}, \bibinfo{year}{2016}.
\newblock \bibinfo{title}{Deep residual learning for image recognition}, in: \bibinfo{booktitle}{Proceedings of the IEEE conference on computer vision and pattern recognition}, pp. \bibinfo{pages}{770--778}.
\bibitem[{Jiang et~al.(2021)Jiang, Han and Bai}]{jiang2021building}
\bibinfo{author}{Jiang, Y.}, \bibinfo{author}{Han, S.}, \bibinfo{author}{Bai, Y.}, \bibinfo{year}{2021}.
\newblock \bibinfo{title}{Building and infrastructure defect detection and visualization using drone and deep learning technologies}.
\newblock \bibinfo{journal}{Journal of Performance of Constructed Facilities} \bibinfo{volume}{35}, \bibinfo{pages}{04021092}.
\newblock \URLprefix \url{https://doi.org/10.1061/(ASCE)CF.1943-5509.0001652}.
\bibitem[{Jing et~al.(2022)Jing, Sheil and Acikgoz}]{jing2022segmentation}
\bibinfo{author}{Jing, Y.}, \bibinfo{author}{Sheil, B.}, \bibinfo{author}{Acikgoz, S.}, \bibinfo{year}{2022}.
\newblock \bibinfo{title}{Segmentation of large-scale masonry arch bridge point clouds with a synthetic simulator and the bridgenet neural network}.
\newblock \bibinfo{journal}{Automation in Construction} \bibinfo{volume}{142}, \bibinfo{pages}{104459}.
\newblock \URLprefix \url{https://doi.org/10.1016/j.autcon.2022.104459}.
\bibitem[{Jing et~al.(2023)Jing, Sheil and Acikgoz}]{jing2023method}
\bibinfo{author}{Jing, Y.}, \bibinfo{author}{Sheil, B.}, \bibinfo{author}{Acikgoz, S.}, \bibinfo{year}{2023}.
\newblock \bibinfo{title}{A method to generate realistic synthetic point clouds of damaged single-span masonry arch bridges}, in: \bibinfo{booktitle}{International Conference on Structural Analysis of Historical Constructions}, \bibinfo{organization}{Springer}. pp. \bibinfo{pages}{436--448}.
\newblock \URLprefix \url{https://doi.org/10.1007/978-3-031-39603-8_36}.
\bibitem[{Jing et~al.(2024a)Jing, Sheil and Acikgoz}]{jing2024lightweight}
\bibinfo{author}{Jing, Y.}, \bibinfo{author}{Sheil, B.}, \bibinfo{author}{Acikgoz, S.}, \bibinfo{year}{2024}a.
\newblock \bibinfo{title}{A lightweight transformer-based neural network for large-scale masonry arch bridge point cloud segmentation}.
\newblock \bibinfo{journal}{Computer-Aided Civil and Infrastructure Engineering} \URLprefix \url{https://doi.org/10.1111/mice.13201}.
\bibitem[{Jing et~al.(2024b)Jing, Zhong, Sheil and Acikgoz}]{jing4819836anomaly}
\bibinfo{author}{Jing, Y.}, \bibinfo{author}{Zhong, J.X.}, \bibinfo{author}{Sheil, B.}, \bibinfo{author}{Acikgoz, S.}, \bibinfo{year}{2024}b.
\newblock \bibinfo{title}{Anomaly detection of cracks in synthetic masonry arch bridge point clouds using fast point feature histograms and patchcore}.
\newblock \bibinfo{journal}{Automation in Construction} \bibinfo{volume}{168}, \bibinfo{pages}{105766}.
\bibitem[{Lee et~al.(2023)Lee, Fan and Sencer}]{lee2023new}
\bibinfo{author}{Lee, E.T.}, \bibinfo{author}{Fan, Z.}, \bibinfo{author}{Sencer, B.}, \bibinfo{year}{2023}.
\newblock \bibinfo{title}{A new approach to detect surface defects from 3d point cloud data with surface normal gabor filter (sngf)}.
\newblock \bibinfo{journal}{Journal of Manufacturing Processes} \bibinfo{volume}{92}, \bibinfo{pages}{196--205}.
\newblock \URLprefix \url{https://doi.org/10.1016/j.jmapro.2023.02.047}.
\bibitem[{Li et~al.(2020)Li, Ma, He, Ren and Liu}]{li2020automatic}
\bibinfo{author}{Li, G.}, \bibinfo{author}{Ma, B.}, \bibinfo{author}{He, S.}, \bibinfo{author}{Ren, X.}, \bibinfo{author}{Liu, Q.}, \bibinfo{year}{2020}.
\newblock \bibinfo{title}{Automatic tunnel crack detection based on u-net and a convolutional neural network with alternately updated clique}.
\newblock \bibinfo{journal}{Sensors} \bibinfo{volume}{20}, \bibinfo{pages}{717}.
\newblock \URLprefix \url{https://doi.org/10.3390/s20030717}.
\bibitem[{Li et~al.(2024a)Li, Rui, Zhu, Lu and Li}]{li2024comprehensive}
\bibinfo{author}{Li, T.}, \bibinfo{author}{Rui, Y.}, \bibinfo{author}{Zhu, H.}, \bibinfo{author}{Lu, L.}, \bibinfo{author}{Li, X.}, \bibinfo{year}{2024}a.
\newblock \bibinfo{title}{Comprehensive digital twin for infrastructure: A novel ontology and graph-based modelling paradigm}.
\newblock \bibinfo{journal}{Advanced Engineering Informatics} \bibinfo{volume}{62}, \bibinfo{pages}{102747}.
\newblock \URLprefix \url{https://doi.org/10.1016/j.aei.2024.102747}.
\bibitem[{Li et~al.(2024b)Li, Yao, Chen, Zhang, Sun, Qian and Wu}]{li2024collaborative}
\bibinfo{author}{Li, Y.}, \bibinfo{author}{Yao, J.}, \bibinfo{author}{Chen, K.}, \bibinfo{author}{Zhang, H.}, \bibinfo{author}{Sun, X.}, \bibinfo{author}{Qian, Q.}, \bibinfo{author}{Wu, X.}, \bibinfo{year}{2024}b.
\newblock \bibinfo{title}{A collaborative anomaly localization method based on multi-modal images}, in: \bibinfo{booktitle}{2024 27th International Conference on Computer Supported Cooperative Work in Design (CSCWD)}, \bibinfo{organization}{IEEE}. pp. \bibinfo{pages}{1322--1327}.
\newblock \URLprefix \url{10.1109/CSCWD61410.2024.10580587}.
\bibitem[{Liao et~al.(2022a)Liao, Yue, Zhang, Tu, Cao, Zou and Li}]{9678126}
\bibinfo{author}{Liao, J.}, \bibinfo{author}{Yue, Y.}, \bibinfo{author}{Zhang, D.}, \bibinfo{author}{Tu, W.}, \bibinfo{author}{Cao, R.}, \bibinfo{author}{Zou, Q.}, \bibinfo{author}{Li, Q.}, \bibinfo{year}{2022}a.
\newblock \bibinfo{title}{Automatic tunnel crack inspection using an efficient mobile imaging module and a lightweight cnn}.
\newblock \bibinfo{journal}{IEEE Transactions on Intelligent Transportation Systems} \bibinfo{volume}{23}, \bibinfo{pages}{15190--15203}.
\newblock \URLprefix \url{10.1109/TITS.2021.3138428}.
\bibitem[{Liao et~al.(2022b)Liao, Yue, Zhang, Tu, Cao, Zou and Li}]{liao2022automatic}
\bibinfo{author}{Liao, J.}, \bibinfo{author}{Yue, Y.}, \bibinfo{author}{Zhang, D.}, \bibinfo{author}{Tu, W.}, \bibinfo{author}{Cao, R.}, \bibinfo{author}{Zou, Q.}, \bibinfo{author}{Li, Q.}, \bibinfo{year}{2022}b.
\newblock \bibinfo{title}{Automatic tunnel crack inspection using an efficient mobile imaging module and a lightweight cnn}.
\newblock \bibinfo{journal}{IEEE Transactions on Intelligent Transportation Systems} \bibinfo{volume}{23}, \bibinfo{pages}{15190--15203}.
\bibitem[{Lin et~al.(2017)Lin, Goyal, Girshick, He and Dollar}]{Lin_2017_ICCV}
\bibinfo{author}{Lin, T.Y.}, \bibinfo{author}{Goyal, P.}, \bibinfo{author}{Girshick, R.}, \bibinfo{author}{He, K.}, \bibinfo{author}{Dollar, P.}, \bibinfo{year}{2017}.
\newblock \bibinfo{title}{Focal loss for dense object detection}, in: \bibinfo{booktitle}{Proceedings of the IEEE International Conference on Computer Vision (ICCV)}.
\newblock \URLprefix \url{https://doi.org/10.48550/arXiv.1708.02002}.
\bibitem[{Lin et~al.(2023)Lin, Li, Xie, Cao and Zhang}]{lin_novel_2023}
\bibinfo{author}{Lin, W.}, \bibinfo{author}{Li, P.}, \bibinfo{author}{Xie, X.}, \bibinfo{author}{Cao, Y.}, \bibinfo{author}{Zhang, Y.}, \bibinfo{year}{2023}.
\newblock \bibinfo{title}{A novel back-analysis approach for the external loads on shield tunnel lining in service based on monitored deformation}.
\newblock \bibinfo{journal}{Structural Control and Health Monitoring} \bibinfo{volume}{2023}, \bibinfo{pages}{8128701}.
\newblock \DOIprefix\doi{10.1155/2023/8128701}.
\bibitem[{Lin et~al.(2024)Lin, Sheil, Zhang, Zhou, Wang and Xie}]{lin_seg2tunnel_2024}
\bibinfo{author}{Lin, W.}, \bibinfo{author}{Sheil, B.}, \bibinfo{author}{Zhang, P.}, \bibinfo{author}{Zhou, B.}, \bibinfo{author}{Wang, C.}, \bibinfo{author}{Xie, X.}, \bibinfo{year}{2024}.
\newblock \bibinfo{title}{Seg2tunnel: A hierarchical point cloud dataset and benchmarks for segmentation of segmental tunnel linings}.
\newblock \bibinfo{journal}{Tunnelling and Underground Space Technology} \bibinfo{volume}{147}, \bibinfo{pages}{105735}.
\newblock \DOIprefix\doi{10.1016/j.tust.2024.105735}.
\bibitem[{Liu et~al.(2024)Liu, Xie, Chen, Li, Wang, Liu, Wang and Zheng}]{liu2024real3d}
\bibinfo{author}{Liu, J.}, \bibinfo{author}{Xie, G.}, \bibinfo{author}{Chen, R.}, \bibinfo{author}{Li, X.}, \bibinfo{author}{Wang, J.}, \bibinfo{author}{Liu, Y.}, \bibinfo{author}{Wang, C.}, \bibinfo{author}{Zheng, F.}, \bibinfo{year}{2024}.
\newblock \bibinfo{title}{Real3d-ad: A dataset of point cloud anomaly detection}.
\newblock \bibinfo{journal}{Advances in Neural Information Processing Systems} \bibinfo{volume}{36}.
\newblock \URLprefix \url{https://doi.org/10.48550/arXiv.2309.13226}.
\bibitem[{Liu et~al.(2011)Liu, Chen and Hauser}]{liu2011lidar}
\bibinfo{author}{Liu, W.}, \bibinfo{author}{Chen, S.}, \bibinfo{author}{Hauser, E.}, \bibinfo{year}{2011}.
\newblock \bibinfo{title}{Lidar-based bridge structure defect detection}.
\newblock \bibinfo{journal}{Experimental Techniques} \bibinfo{volume}{35}, \bibinfo{pages}{27--34}.
\newblock \URLprefix \url{https://doi.org/10.1111/j.1747-1567.2010.00644.x}.
\bibitem[{Liu et~al.(2019)Liu, Cao, Wang and Wang}]{liu2019computer}
\bibinfo{author}{Liu, Z.}, \bibinfo{author}{Cao, Y.}, \bibinfo{author}{Wang, Y.}, \bibinfo{author}{Wang, W.}, \bibinfo{year}{2019}.
\newblock \bibinfo{title}{Computer vision-based concrete crack detection using u-net fully convolutional networks}.
\newblock \bibinfo{journal}{Automation in Construction} \bibinfo{volume}{104}, \bibinfo{pages}{129--139}.
\newblock \URLprefix \url{https://doi.org/10.1016/j.autcon.2019.04.005}.
\bibitem[{Logoglu et~al.(2016)Logoglu, Kalkan and Temizel}]{logoglu2016cospair}
\bibinfo{author}{Logoglu, K.B.}, \bibinfo{author}{Kalkan, S.}, \bibinfo{author}{Temizel, A.}, \bibinfo{year}{2016}.
\newblock \bibinfo{title}{Cospair: colored histograms of spatial concentric surflet-pairs for 3d object recognition}.
\newblock \bibinfo{journal}{Robotics and Autonomous Systems} \bibinfo{volume}{75}, \bibinfo{pages}{558--570}.
\newblock \URLprefix \url{https://doi.org/10.1016/j.robot.2015.09.027}.
\bibitem[{Loverdos and Sarhosis(2022)}]{loverdos2022automatic}
\bibinfo{author}{Loverdos, D.}, \bibinfo{author}{Sarhosis, V.}, \bibinfo{year}{2022}.
\newblock \bibinfo{title}{Automatic image-based brick segmentation and crack detection of masonry walls using machine learning}.
\newblock \bibinfo{journal}{Automation in Construction} \bibinfo{volume}{140}, \bibinfo{pages}{104389}.
\newblock \URLprefix \url{https://doi.org/10.1016/j.autcon.2022.104389}.
\bibitem[{Mohammadi et~al.(2019)Mohammadi, Wood and Wittich}]{mohammadi2019non}
\bibinfo{author}{Mohammadi, M.E.}, \bibinfo{author}{Wood, R.L.}, \bibinfo{author}{Wittich, C.E.}, \bibinfo{year}{2019}.
\newblock \bibinfo{title}{Non-temporal point cloud analysis for surface damage in civil structures}.
\newblock \bibinfo{journal}{ISPRS International Journal of Geo-Information} \bibinfo{volume}{8}, \bibinfo{pages}{527}.
\newblock \URLprefix \url{https://doi.org/10.3390/ijgi8120527}.
\bibitem[{Narazaki et~al.(2021)Narazaki, Hoskere, Yoshida, Spencer and Fujino}]{narazaki2021synthetic}
\bibinfo{author}{Narazaki, Y.}, \bibinfo{author}{Hoskere, V.}, \bibinfo{author}{Yoshida, K.}, \bibinfo{author}{Spencer, B.F.}, \bibinfo{author}{Fujino, Y.}, \bibinfo{year}{2021}.
\newblock \bibinfo{title}{Synthetic environments for vision-based structural condition assessment of japanese high-speed railway viaducts}.
\newblock \bibinfo{journal}{Mechanical Systems and Signal Processing} \bibinfo{volume}{160}, \bibinfo{pages}{107850}.
\newblock \URLprefix \url{https://doi.org/10.1016/j.ymssp.2021.107850}.
\bibitem[{Nguyen et~al.(2019)Nguyen, Lou, Klar and Brox}]{nguyen2019anomaly}
\bibinfo{author}{Nguyen, D.T.}, \bibinfo{author}{Lou, Z.}, \bibinfo{author}{Klar, M.}, \bibinfo{author}{Brox, T.}, \bibinfo{year}{2019}.
\newblock \bibinfo{title}{Anomaly detection with multiple-hypotheses predictions}, in: \bibinfo{booktitle}{International Conference on Machine Learning}, \bibinfo{organization}{PMLR}. pp. \bibinfo{pages}{4800--4809}.
\newblock \URLprefix \url{https://doi.org/10.48550/arXiv.1810.13292}.
\bibitem[{Perez et~al.(2019)Perez, Tah and Mosavi}]{perez2019deep}
\bibinfo{author}{Perez, H.}, \bibinfo{author}{Tah, J.H.}, \bibinfo{author}{Mosavi, A.}, \bibinfo{year}{2019}.
\newblock \bibinfo{title}{Deep learning for detecting building defects using convolutional neural networks}.
\newblock \bibinfo{journal}{Sensors} \bibinfo{volume}{19}, \bibinfo{pages}{3556}.
\newblock \URLprefix \url{https://doi.org/10.3390/s19163556}.
\bibitem[{Pesci et~al.(2011)Pesci, Casula and Boschi}]{pesci2011laser}
\bibinfo{author}{Pesci, A.}, \bibinfo{author}{Casula, G.}, \bibinfo{author}{Boschi, E.}, \bibinfo{year}{2011}.
\newblock \bibinfo{title}{Laser scanning the garisenda and asinelli towers in bologna (italy): Detailed deformation patterns of two ancient leaning buildings}.
\newblock \bibinfo{journal}{Journal of cultural heritage} \bibinfo{volume}{12}, \bibinfo{pages}{117--127}.
\newblock \URLprefix \url{https://doi.org/10.1016/j.culher.2011.01.002}.
\bibitem[{Protopapadakis et~al.(2019)Protopapadakis, Voulodimos, Doulamis, Doulamis and Stathaki}]{protopapadakis2019automatic}
\bibinfo{author}{Protopapadakis, E.}, \bibinfo{author}{Voulodimos, A.}, \bibinfo{author}{Doulamis, A.}, \bibinfo{author}{Doulamis, N.}, \bibinfo{author}{Stathaki, T.}, \bibinfo{year}{2019}.
\newblock \bibinfo{title}{Automatic crack detection for tunnel inspection using deep learning and heuristic image post-processing}.
\newblock \bibinfo{journal}{Applied intelligence} \bibinfo{volume}{49}, \bibinfo{pages}{2793--2806}.
\newblock \URLprefix \url{https://doi.org/10.1007/s10489-018-01396-y}.
\bibitem[{Reitmann et~al.(2021)Reitmann, Neumann and Jung}]{reitmann2021blainder}
\bibinfo{author}{Reitmann, S.}, \bibinfo{author}{Neumann, L.}, \bibinfo{author}{Jung, B.}, \bibinfo{year}{2021}.
\newblock \bibinfo{title}{Blainder—a blender ai add-on for generation of semantically labeled depth-sensing data}.
\newblock \bibinfo{journal}{Sensors} \bibinfo{volume}{21}, \bibinfo{pages}{2144}.
\newblock \URLprefix \url{https://doi.org/10.3390/s21062144}.
\bibitem[{del R{\'\i}o-Barral et~al.(2022)del R{\'\i}o-Barral, Soil{\'a}n, Gonz{\'a}lez-Collazo and Arias}]{del2022pavement}
\bibinfo{author}{del R{\'\i}o-Barral, P.}, \bibinfo{author}{Soil{\'a}n, M.}, \bibinfo{author}{Gonz{\'a}lez-Collazo, S.M.}, \bibinfo{author}{Arias, P.}, \bibinfo{year}{2022}.
\newblock \bibinfo{title}{Pavement crack detection and clustering via region-growing algorithm from 3d mls point clouds}.
\newblock \bibinfo{journal}{Remote Sensing} \bibinfo{volume}{14}, \bibinfo{pages}{5866}.
\newblock \URLprefix \url{https://doi.org/10.3390/rs14225866}.
\bibitem[{Roth et~al.(2022)Roth, Pemula, Zepeda, Sch{\"o}lkopf, Brox and Gehler}]{roth2022towards}
\bibinfo{author}{Roth, K.}, \bibinfo{author}{Pemula, L.}, \bibinfo{author}{Zepeda, J.}, \bibinfo{author}{Sch{\"o}lkopf, B.}, \bibinfo{author}{Brox, T.}, \bibinfo{author}{Gehler, P.}, \bibinfo{year}{2022}.
\newblock \bibinfo{title}{Towards total recall in industrial anomaly detection}, in: \bibinfo{booktitle}{Proceedings of the IEEE/CVF Conference on Computer Vision and Pattern Recognition}, pp. \bibinfo{pages}{14318--14328}.
\newblock \URLprefix \url{10.1109/CVPR52688.2022.01392}.
\bibitem[{Rusu et~al.(2009)Rusu, Blodow and Beetz}]{rusu2009fast}
\bibinfo{author}{Rusu, R.B.}, \bibinfo{author}{Blodow, N.}, \bibinfo{author}{Beetz, M.}, \bibinfo{year}{2009}.
\newblock \bibinfo{title}{Fast point feature histograms (fpfh) for 3d registration}, in: \bibinfo{booktitle}{2009 IEEE international conference on robotics and automation}, \bibinfo{organization}{IEEE}. pp. \bibinfo{pages}{3212--3217}.
\newblock \URLprefix \url{10.1109/ROBOT.2009.5152473}.
\bibitem[{Schlegl et~al.(2019)Schlegl, Seeb{\"o}ck, Waldstein, Langs and Schmidt-Erfurth}]{schlegl2019f}
\bibinfo{author}{Schlegl, T.}, \bibinfo{author}{Seeb{\"o}ck, P.}, \bibinfo{author}{Waldstein, S.M.}, \bibinfo{author}{Langs, G.}, \bibinfo{author}{Schmidt-Erfurth, U.}, \bibinfo{year}{2019}.
\newblock \bibinfo{title}{f-anogan: Fast unsupervised anomaly detection with generative adversarial networks}.
\newblock \bibinfo{journal}{Medical image analysis} \bibinfo{volume}{54}, \bibinfo{pages}{30--44}.
\newblock \URLprefix \url{https://doi.org/10.1016/j.media.2019.01.010}.
\bibitem[{Sener and Savarese(2017)}]{sener2017active}
\bibinfo{author}{Sener, O.}, \bibinfo{author}{Savarese, S.}, \bibinfo{year}{2017}.
\newblock \bibinfo{title}{Active learning for convolutional neural networks: A core-set approach}.
\newblock \bibinfo{journal}{arXiv preprint arXiv:1708.00489} \URLprefix \url{https://doi.org/10.48550/arXiv.1708.00489}.
\bibitem[{Sta{\l}owska et~al.(2022)Sta{\l}owska, Suchocki and Rutkowska}]{stalowska2022crack}
\bibinfo{author}{Sta{\l}owska, P.}, \bibinfo{author}{Suchocki, C.}, \bibinfo{author}{Rutkowska, M.}, \bibinfo{year}{2022}.
\newblock \bibinfo{title}{Crack detection in building walls based on geometric and radiometric point cloud information}.
\newblock \bibinfo{journal}{Automation in Construction} \bibinfo{volume}{134}, \bibinfo{pages}{104065}.
\newblock \URLprefix \url{https://doi.org/10.1016/j.autcon.2021.104065}.
\bibitem[{Vaswani et~al.(2017)Vaswani, Shazeer, Parmar, Uszkoreit, Jones, Gomez, Kaiser and Polosukhin}]{vaswani2017attention}
\bibinfo{author}{Vaswani, A.}, \bibinfo{author}{Shazeer, N.}, \bibinfo{author}{Parmar, N.}, \bibinfo{author}{Uszkoreit, J.}, \bibinfo{author}{Jones, L.}, \bibinfo{author}{Gomez, A.N.}, \bibinfo{author}{Kaiser, {\L}.}, \bibinfo{author}{Polosukhin, I.}, \bibinfo{year}{2017}.
\newblock \bibinfo{title}{Attention is all you need}.
\newblock \bibinfo{journal}{Advances in neural information processing systems} \bibinfo{volume}{30}.
\newblock \URLprefix \url{10.5555/3295222.3295349}.
\bibitem[{Wan et~al.(2021)Wan, Gao, Li and Wen}]{wan2021industrial}
\bibinfo{author}{Wan, Q.}, \bibinfo{author}{Gao, L.}, \bibinfo{author}{Li, X.}, \bibinfo{author}{Wen, L.}, \bibinfo{year}{2021}.
\newblock \bibinfo{title}{Industrial image anomaly localization based on gaussian clustering of pretrained feature}.
\newblock \bibinfo{journal}{IEEE Transactions on Industrial Electronics} \bibinfo{volume}{69}, \bibinfo{pages}{6182--6192}.
\newblock \URLprefix \url{10.1109/TIE.2021.3094452}.
\bibitem[{Wang et~al.(2023)Wang, Peng, Zhang, Yi, Wang and Wang}]{wang2023multimodal}
\bibinfo{author}{Wang, Y.}, \bibinfo{author}{Peng, J.}, \bibinfo{author}{Zhang, J.}, \bibinfo{author}{Yi, R.}, \bibinfo{author}{Wang, Y.}, \bibinfo{author}{Wang, C.}, \bibinfo{year}{2023}.
\newblock \bibinfo{title}{Multimodal industrial anomaly detection via hybrid fusion}, in: \bibinfo{booktitle}{Proceedings of the IEEE/CVF Conference on Computer Vision and Pattern Recognition}, pp. \bibinfo{pages}{8032--8041}.
\newblock \URLprefix \url{10.1109/CVPR52729.2023.00776}.
\bibitem[{Weglarczyk(2018)}]{wkeglarczyk2018kernel}
\bibinfo{author}{Weglarczyk, S.}, \bibinfo{year}{2018}.
\newblock \bibinfo{title}{Kernel density estimation and its application}, in: \bibinfo{booktitle}{ITM web of conferences}, \bibinfo{organization}{EDP Sciences}. p. \bibinfo{pages}{00037}.
\bibitem[{Ye et~al.(2018)Ye, Acikgoz, Pendrigh, Riley and DeJong}]{ye2018mapping}
\bibinfo{author}{Ye, C.}, \bibinfo{author}{Acikgoz, S.}, \bibinfo{author}{Pendrigh, S.}, \bibinfo{author}{Riley, E.}, \bibinfo{author}{DeJong, M.}, \bibinfo{year}{2018}.
\newblock \bibinfo{title}{Mapping deformations and inferring movements of masonry arch bridges using point cloud data}.
\newblock \bibinfo{journal}{Engineering Structures} \bibinfo{volume}{173}, \bibinfo{pages}{530--545}.
\newblock \URLprefix \url{https://doi.org/10.1016/j.engstruct.2018.06.094}.
\bibitem[{Yi and Yoon(2020)}]{yi2020patch}
\bibinfo{author}{Yi, J.}, \bibinfo{author}{Yoon, S.}, \bibinfo{year}{2020}.
\newblock \bibinfo{title}{Patch svdd: Patch-level svdd for anomaly detection and segmentation}, in: \bibinfo{booktitle}{Proceedings of the Asian conference on computer vision}.
\newblock \URLprefix \url{https://doi.org/10.1007/978-3-030-69544-6_23}.

\end{thebibliography}

\end{document}